\documentclass[11pt]{scaleai-paper}

\usepackage{amsmath}
\usepackage{amsfonts}
\usepackage{amssymb}
\usepackage{amsthm}
\usepackage{booktabs}
\usepackage{tabularx}
\usepackage{tabulary}
\usepackage{multirow}
\usepackage{subcaption}
\usepackage{float}
\usepackage{threeparttable}
\usepackage[square,numbers,sort&compress]{natbib}
\usepackage{xspace}
\usepackage{url}
\usepackage[colorlinks=true,linkcolor=scaleLink,citecolor=scaleLink,urlcolor=scaleLink]{hyperref}
\usepackage[capitalise,nameinlink]{cleveref}
\usepackage{tikz}
\usetikzlibrary{arrows.meta, positioning, fit, decorations.pathreplacing}
\newcommand{\attendcell}[2]{\fill[scaleBlue!55] (#1,#2) rectangle ++(1.15,-1.15); \draw[scaleGray!40] (#1,#2) rectangle ++(1.15,-1.15);}
\newcommand{\maskcell}[2]{\draw[fill=white,draw=scaleGray!40] (#1,#2) rectangle ++(1.15,-1.15);}
\newcommand{\causalcell}[2]{\draw[draw=scaleGray!40] (#1,#2) rectangle ++(1.15,-1.15); \fill[scaleBlue!55] (#1,#2) -- ++(0,-1.15) -- ++(1.15,0) -- cycle;}

\newcolumntype{Y}{>{\RaggedRight\arraybackslash}X}
\let\svthefootnote\thefootnote
\newcommand\freefootnote[1]{%
  \let\thefootnote\relax%
  \footnotetext{#1}%
  \let\thefootnote\svthefootnote%
}

\graphicspath{{figures/}}

\papertype{Scale AI Research}
\secondlogopath{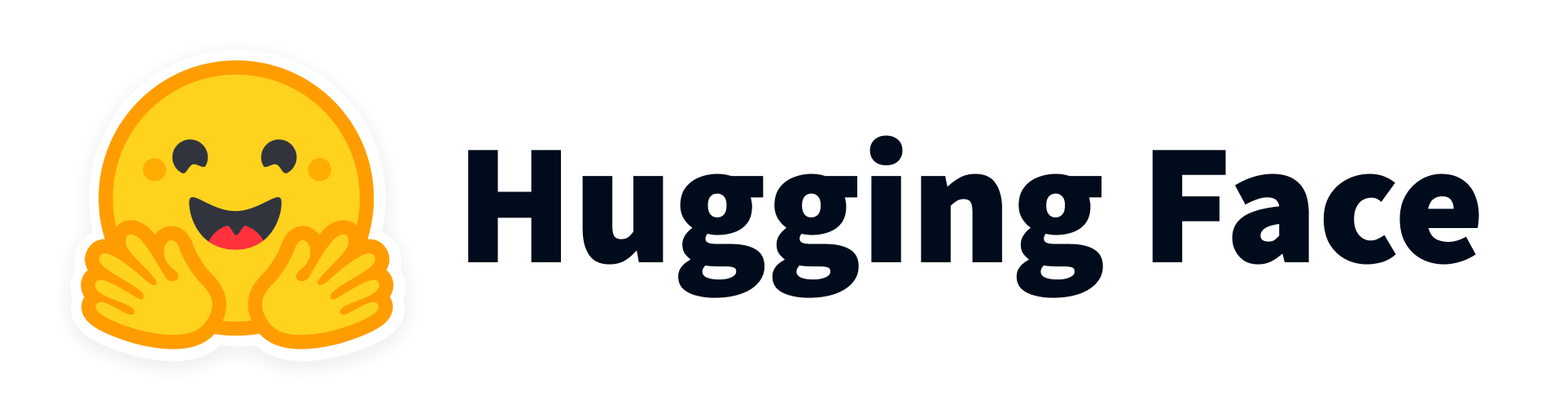}
\secondlogoheight{0.32in}
\headerlogoheight{0.22in}
\titlefontsize{19}{23}

\title{FineART: Fine-Grained Annotated Robotic Trajectory Dataset and Vision-Language-Action Model for Bimanual Manipulation}

\author[1,*]{Jade Choghari}
\author[2,*]{Pepijn Kooijmans}
\author[1]{\\Mansi Agarwal}
\author[1,3,4]{Yusuf Umut Ciftci}
\author[1]{Aseem Doriwala}
\author[1]{\\Catherine Weaver}
\author[\textdaggerdbl]{Mouli Sivapurapu}
\author[1]{Kai Yang}
\author[2,\textdagger]{\\Thomas Wolf}
\author[1,\textdagger]{Jackson Lee}
\author[1,\textdagger]{Pragna~Mannam}

\affil[1]{Scale AI}
\affil[2]{Hugging Face}
\affil[3]{University of Southern California}
\affil[4]{Stanford University}

\begin{document}

\freefootnote{*Equal contribution. \quad \textdagger\ Equal senior contribution. \quad \textdaggerdbl\ Work done while at Scale AI.}

\maketitle

\begin{abstract}
Robots operating in real-world environments must often execute complex, multi-step bimanual tasks over long horizons rather than single, isolated actions. Current manipulation datasets provide limited support for this capability: although single-arm datasets reach hundreds of thousands of trajectories, they typically provide only one high-level instruction per episode, while existing bimanual datasets provide subtask annotations for only part of their recorded hours. We present FineART, a densely annotated bimanual manipulation dataset comprising $40{,}543$ episodes ($1{,}718$ hours) and $533{,}913$ subtasks across $151$ tasks. We also introduce FineART-VLA, a vision-language-action policy that predicts its own next subtask to guide its actions. Mid-training on FineART's subtask annotations improves FineART-VLA's success at following spatial instructions from $32.0\%$ to $100.0\%$. With step-by-step human subtask guidance, success on unseen long-horizon tasks increases from $16.0\%$ to $76.0\%$. Furthermore, FineART-VLA matches baseline performance on a new robot with $10\times$ less fine-tuning data and generalizes zero-shot to unseen tasks. We open-source the full dataset, model weights, and training code.
\end{abstract}

\enlargethispage{2\baselineskip}
\begin{figure}[H]
\centering
\includegraphics[width=0.98\linewidth]{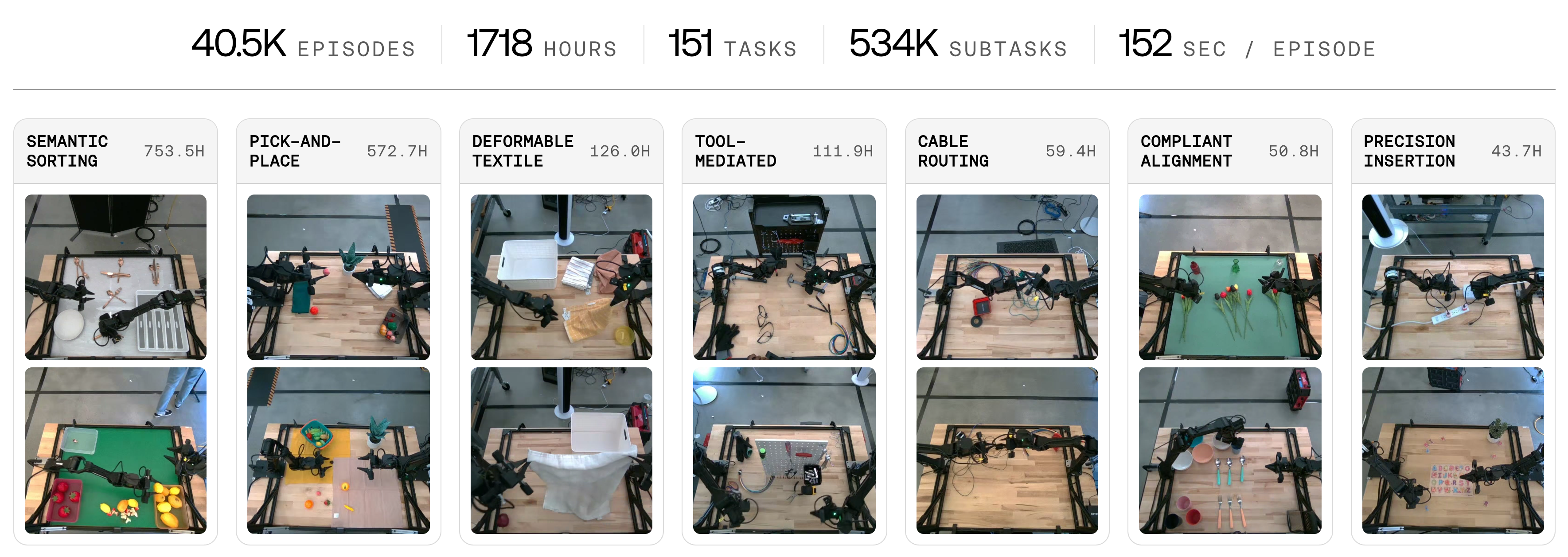}
\caption{Representative demonstrations from FineART dataset. FineART is the largest subtask-annotated manipulation dataset to date, consisting of 40.5K episodes and 534K subtasks, totaling 1,718 hours.}
\label{fig:teaser}
\end{figure}

\section{Introduction}
\label{sec:introduction}

\begin{quote}
\textit{``Divide each difficulty into as many parts as is feasible and necessary to resolve it.''}
\begin{flushright}
\footnotesize --- Ren\'e Descartes, \textit{Discourse on Method} (1637), Part II
\end{flushright}
\end{quote}

Robotic assistants that work in real households and workplaces must handle long-horizon tasks -- such as preparing a meal, sorting tools, or packing a bag -- that demand dozens of coordinated two-handed actions over several minutes. Humans manage this complexity by continuously breaking a task into a running sequence of subtasks (e.g., pick up plate, pick up toast, put toast on plate) instead of holding one instruction for the full duration. Giving a robot policy that same ability requires two foundations: (1) long-horizon bimanual manipulation data and (2) language supervision that marks subtask boundaries densely and persistently enough to train on directly.

While single-arm datasets \cite{oxe, droid} scale to hundreds of thousands of trajectories, they consist mainly of short pick-and-place episodes associated with single instructions. Recent bimanual datasets \cite{aloha, mobilealoha, rh20t} address embodiment gaps and scale, but even the largest \cite{abc} only annotates subtask labels for a fraction of its hours. Furthermore, they report diversity in aggregate, obscuring whether they cover contact-rich, long-tail skills like insertion, deformable object manipulation, or tool use. Existing annotation pipelines \cite{calvin, saycan, rth} make this worse: they either assign broad, high-level labels to entire videos or try to generate subtasks on the fly during execution, rather than giving the model step-by-step guidance directly throughout training.

Without step-by-step labels during training, current models struggle to handle long tasks. Flat vision-language-action models \cite{rt2, openvla} map an instruction and image directly to actions, with no way to track progress through a multi-minute task, while hierarchical alternatives \cite{saycan, hirobot} split planning and execution across two separate models. No existing dataset-and-model combination offers persistent, dense subtask training for long-horizon bimanual manipulation inside a single end-to-end policy. FineART closes this gap: a bimanual manipulation dataset where every one of its $1{,}718$ hours across $40{,}543$ episodes (averaging $152.5$ seconds each) carries dense subtask annotations -- $533{,}913$ subtask labels in total, roughly $13$ per episode -- paired with a training recipe that treats these subtasks as persistent supervision inside a single end-to-end policy.

Our contributions are:
\begin{itemize}[itemsep=0pt, topsep=2pt, parsep=0pt]
    \item \textbf{FineART Dataset.} An open-source bimanual manipulation dataset with dense subtask annotations, containing $40{,}543$ labeled episodes and $533{,}913$ subtask labels.
    \item \textbf{FineART-VLA.} A single end-to-end policy, extended from a pretrained vision-language-action backbone, that pairs System-2 subtask prediction in language with System-1 continuous-action execution using hierarchical inference and knowledge insulation instead of splitting planning and execution across two separate models.
    \item \textbf{Dense Subtask Training.} Dense subtask training improves spatial grounding and long-horizon instruction following, raising success rate from $32.0\%$ to $100.0\%$ on a spatial disambiguation task and from $16.0\%$ to $76.0\%$ on a long-horizon task requiring human corrections, compared to a matched policy trained without subtask training.
    \item \textbf{Cross-Embodiment Mid-Training.} Mid-training on FineART enables zero-shot transfer to unseen tasks. After fine-tuning on only five atomic tasks, the policy generalizes to unseen, multi-stage tasks on a distinct robot platform (YAM), reaching $28.0\%$ success on prepare breakfast, $12.0\%$ on sort tools, and $60.0\%$ on put cable into bin under distractors.
\end{itemize}
\enlargethispage{1\baselineskip}

We open-source the model and training code alongside the dataset.

\section{Related Work}
\label{sec:related-work}

\subsection{Large-Scale Robot Learning Datasets}
Robot learning has scaled in waves. An early single-arm dataset \cite{mtopt} established teleoperated data collection at growing scale, but it stayed within a handful of tasks and seconds-long episodes. Later efforts \cite{oxe, droid} pushed scale further still, aggregating over a million and 76k trajectories respectively across dozens of platforms and hundreds of in-the-wild scenes, yet both remain dominated by single-arm, short-horizon pick-and-place, with one instruction typically covering an entire episode. Bimanual datasets close part of that gap \cite{aloha, mobilealoha, rh20t, aistbimanip}: two introduce low-cost teleoperation for contact-rich two-arm manipulation, one adds force and audio modalities across 110k sequences, and the last extends coverage to over 100 more tasks. Two very recent efforts push bimanual scale further still: MolmoAct2's Bimanual YAM release \cite{molmoact2} and the ABC-130K dataset \cite{abc}, the latter also annotating a 1,552-hour subset with sub-episode task labels. \Cref{tab:dataset-comparison} compares all of these datasets across trajectory count, task count, and annotated hours: FineART is the only dataset combining bimanual manipulation, multi-minute episodes, and dense sub-episode labels across its entire duration, rather than a curated subset of it.

\begin{table}[t]
\centering
\small
\setlength{\tabcolsep}{6pt}
\renewcommand{\arraystretch}{1.15}
\begin{tabular}{@{}lccccc@{}}
\toprule
\textbf{Dataset} & \textbf{Traj.} & \textbf{Tasks} & \textbf{Hours} & \textbf{Subtask Hours} & \textbf{Sec./Ep.} \\
\midrule
MT-Opt \cite{mtopt} & 800,000 & 12 & 5,556 & 0 & 25 \\
RH20T \cite{rh20t} & 110,000 & 147 & 1,111 & 0 & 36.4 \\
RoboSet \cite{roboset} & 7,500 & 38 & N/A & 0 & N/A \\
BridgeData V2 \cite{bridgedata} & 60,096 & 13 & 127 & 0 & 7.6 \\
Open X-Embodiment \cite{oxe} & 1M+ & 500+ & N/A & 0 & N/A \\
DROID \cite{droid} & 76,000 & 86 & 350 & 0 & 16.6 \\
MolmoAct2 (Bimanual YAM) \cite{molmoact2} & 34,500 & 28 & 720 & 0 & 75.1 \\
ABC-130K \cite{abc} & 134,806 & 195 & 3,553 & 1,552 & 94.9 \\
\textbf{FineART (Ours)} & \textbf{40,543} & \textbf{151} & \textbf{1,718} & \textbf{1,718} & \textbf{152.5} \\
\bottomrule
\end{tabular}
\caption{Comparison of open-source robot manipulation datasets. FineART has the most densely annotated subtask hours. N/A: not reported.}
\label{tab:dataset-comparison}
\end{table}

\subsection{Data Curation and Annotation Pipelines}
Dense temporal annotation of long videos originated in captioning work \cite{krishna2017dense}, which jointly localized and described events in multi-minute footage. Robotics applied coarser versions of the same idea \cite{calvin, saycan}, pairing subtasks with play data or decomposing an instruction into skills online rather than annotating the training data itself.
A closer prior effort \cite{rth} inserted an intermediate language-motion layer between task and action, but its hierarchy was produced online rather than as a persistent, dataset-wide label. FineART instead annotates every episode directly, producing $533{,}913$ subtask labels (an average of $13$ per episode) that are available as fixed training targets rather than being generated at inference time.

\subsection{Language-Conditioned Policies}
Augmenting end-to-end policies with an explicit high-level reasoning step improved performance on long-horizon tasks \cite{saycan, hirobot, li2025hamster, geminirobotics2025}, particularly when the high-level step could draw on a pretrained LLM or VLM. Flat vision-language-action models \cite{rt2, openvla} skip this step, mapping a single instruction and image directly to actions; one extension \cite{crossformer} carries the same recipe to heterogeneous embodiments without unifying the action space. These architectures degrade over long horizons because the model does not track task progress. Most hierarchical methods instead split the two roles across separate models, with a VLM predicting semantic subtasks and a distinct low-level policy executing them \cite{saycan, hirobot, li2025hamster}; others \cite{languagetable} conditioned on short natural-language subtasks mined from unstructured play data rather than a planner's output. FineART instead uses a single model for both stages, closer to chain-of-thought \cite{wei2022chain} than to a two-model pipeline, built on a $\pi_{0.5}$ flow-matching backbone \cite{pi5} and knowledge insulation \cite{KI} to prevent continuous-action gradients from destroying pretrained VLM representations. Two concurrent efforts pursued closely related ideas: dense per-clip language annotation for policy learning \cite{demian}, and a diffusion policy conditioned on foundation-model-generated subtasks \cite{sadp}; a third \cite{galaxea} paired a VLM subtask planner with a flow-matching controller in a similar spirit. None reported a dataset combining bimanual multi-minute episodes with persistent sub-episode labels at FineART's scale.

\section{FineART Dataset}
\label{sec:FineART}

FineART is a bimanual manipulation dataset consisting of $40{,}543$ teleoperated episodes ($1{,}718$\,hours, $185.5$\,M frames) spanning $151$ tasks. It is the largest subtask-annotated robot dataset to the best of our knowledge, providing full temporal supervision across $100\%$ of its hours and tasks.

\begin{figure}[t]
\centering
\includegraphics[width=\linewidth]{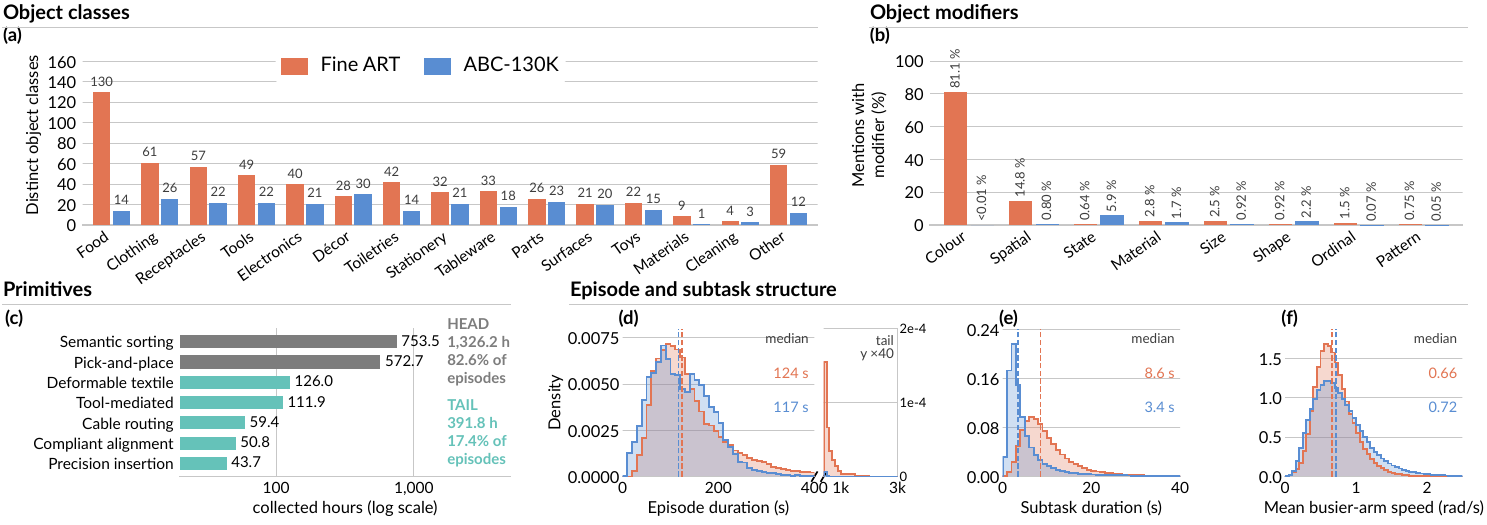}
\caption{FineART dataset overview. (a) Distinct object classes represented in FineART and ABC-130K across semantic categories. (b) Frequency of descriptive object modifiers in subtask annotations. (c) Collected hours across manipulation primitives, highlighting the high-volume head and contact-rich tail. (d--f) Episode duration, subtask duration, and arm-speed distributions, respectively, compared with ABC-130K.}
\label{fig:dataset-overview}
\end{figure}

\subsection{Data Collection Protocol}
\label{sec:protocol}
All data are collected on the Trossen Stationary AI bimanual leader--follower system by human teleoperators. Each follower arm carries an ALOHA2 gripper rated for a $1$\,kg payload and executes a $50$\,Hz control loop, with joint telemetry logged at $300$\,Hz. Visual state is recorded at $30$\,fps ($1280\times720$) across four Intel RealSense D405 cameras: a bird's-eye view for global tabletop context, a worm's-eye view for low-angle contact physics, and dual wrist-mounted cameras for egocentric local manipulation.

A second independent team of human annotators labeled trajectories with natural task descriptions, without VLM assistance. Finally, a third independent team of human auditors performed quality control on a statistically significant sample of the labeled trajectories. This process ensures that the dataset is free of labeling bias from the teleoperators, and that the annotations are accurate, complete, and consistent.

\begin{figure}[t]
\centering
\includegraphics[width=0.6\linewidth]{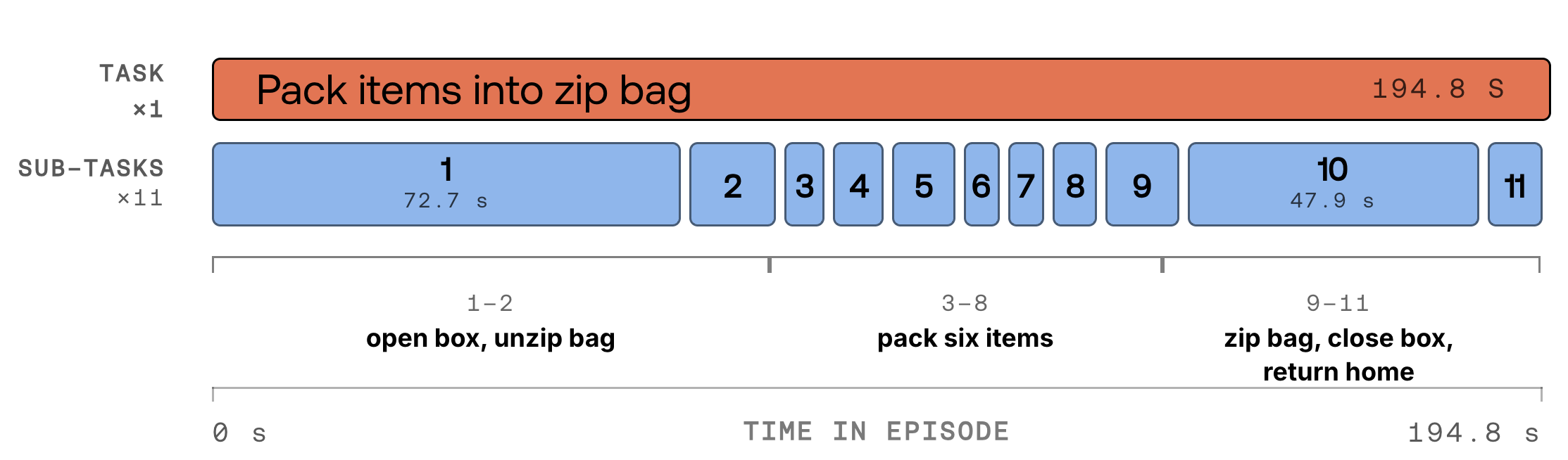}
\caption{FineART annotation schema. A multi-stage task with demonstration-level and temporally segmented subtask annotations. Subtask labels shown are illustrative categorical examples, not the actual annotations.}
\label{fig:schema}
\end{figure}

Labeling yields two granularities (\cref{fig:schema}).
\textbf{Demonstration Label:} Whole-episode text summary paired with metadata describing execution quality and outcome success.
\textbf{Subtask Labels:} Contiguous, non-overlapping temporal segments containing untemplated free-text task descriptions.

\subsection{Dataset Composition: Long-Horizon Multi-Stage Tasks}
\label{sec:composition}

While single-arm datasets like Bridge V2 \cite{bridgedata} and DROID \cite{droid} provide large trajectory volumes, they pair multiminute trajectories with a single instruction. ABC-130K \cite{abc} represents the only prior single-embodiment dataset providing timestamped subtask text alongside raw trajectories. Therefore, we use ABC-130K as a reference while analyzing our dataset composition.

FineART features long-horizon tasks that require multiple manipulation subtasks to complete, averaging 13.17 subtasks per task. The median episode duration is approximately $2.0$\,min, with a long tail of episodes lasting more than $10$\,min (\cref{fig:dataset-overview}d).

\subsubsection{Subtask Composition}
\label{sec:subtasks}
FineART deliberately concentrates most of its subtasks on a small set of reusable manipulation primitives, then adds a long tail of physically distinct, contact-rich skills on top. In a mid-training dataset, a policy needs to see a primitive many times to learn it reliably, while the contact-rich tail tests whether that learned representation actually carries over to harder, fine-grained tasks.

To capture manipulation diversity, we define a primary vocabulary of $31$ leading verbs categorized into seven fundamental manipulation primitives (\cref{fig:dataset-overview}c).

While the majority of the dataset ($1{,}326.2$\,h; $77.2\%$ of hours) concentrates on gross pick-and-place ($572.7$\,h) and semantic organization ($753.5$\,h) to build core spatial manipulation abilities, the remaining $391.8$\,h ($22.8\%$ of hours) forms a long tail of contact-rich operations. This long tail covers tight tolerances, near-unbounded configuration states, and complex contact dynamics across deformable textile manipulation ($126.0$\,h), tool-mediated assembly ($111.9$\,h), flexible linear object routing ($59.4$\,h), compliant alignment ($50.8$\,h), and high-precision mechanical insertion ($43.7$\,h).

Though smaller in total volume, this tail provides the essential demonstration signal needed for policies to move beyond simple prehension toward fine-motor, contact-guided behavior. \Cref{sec:experiments} tests that hypothesis directly by ablating the tail from the training mixture.

We observe that FineART subtasks are about $2.5$ times as long as ABC-130K subtasks, with median duration of $8.6$\,s versus $3.5$\,s (\cref{fig:dataset-overview}e), even though the datasets have similar arm-speed distributions (\cref{fig:dataset-overview}f). This indicates that FineART subtasks capture more multi-stage manipulation (stack, remove, close), while ABC-130K subtasks often segment simpler, individual actions (pass, grab, flatten).

\subsubsection{Object Diversity}
Object diversity is introduced during collection through object rotation and spatial randomization. Every five episodes, a unique set of objects is assigned to each task-operator pair. Objects are repositioned and reoriented for each episode so that no arrangement is repeated. This protocol varies both the objects used to perform tasks and the spatial configurations in which they must be manipulated.

To compare object diversity, we apply a shared rule-based parser to the subtask annotations of FineART and ABC-130K. After spelling normalization and typo correction, we extract noun phrases and represent object classes by singularized head nouns, retaining noun qualifiers for generic heads (e.g., \textit{chess piece}). Preceding modifiers are retained separately as descriptors and tagged using lexicons for color, material, size, shape, pattern, state, spatial position, and ordinal attributes. The parser filters non-object expressions such as regions, robot parts, pronouns, and measurements. These statistics measure objects named in the annotations: a class is a normalized object label, and a mention is an occurrence of that label.

\Cref{fig:dataset-overview}a shows broad coverage across everyday object categories, spanning $657$ distinct object classes in total. FineART contains more distinct labels than ABC-130K in food, clothing, receptacles, tools, electronics, and toiletries. Coverage also extends to stationery, tableware, parts, toys, and materials, with comparable coverage of surfaces and fewer labels for d\'ecor. Thus, the repeated manipulation primitives described above are demonstrated across a range of object identities, materials, and geometries. Because this comparison is annotation-based, it reflects both collection diversity and how specifically annotators name objects.

The \textit{Other} category contains extracted labels that could not be mapped to a semantic category from the noun alone. FineART entries include ambiguous nouns such as \textit{stick} and \textit{mesh}, and parser leftovers such as \textit{form} from the phrase ``to form a loop.'' In ABC-130K, approximately $95\%$ of these mentions use generic labels such as \textit{personal item}, \textit{distractor}, \textit{chemistry item}, and \textit{product}.

\subsubsection{Descriptive Specificity}
The annotations describe not only which object class is involved, but also attributes that distinguish an object within a scene. \Cref{fig:dataset-overview}b reports descriptor frequency per object mention. Color modifiers occur in $81.1\%$ of FineART mentions, compared with less than $0.01\%$ in ABC-130K; spatial modifiers occur in $14.8\%$ versus $0.80\%$. Material, size, ordinal, and pattern descriptors provide additional distinctions. ABC-130K uses state and shape modifiers more frequently, largely because of its many deformable-object demonstrations featuring descriptors such as \textit{folded}, \textit{flattened}, and \textit{crumpled}.

This descriptive detail is reflected in the diversity of the annotation text: FineART contains $110{,}299$ distinct subtask strings and $23{,}589$ distinct object-referring expressions, compared with $7{,}977$ and $534$, respectively, in ABC-130K. In cases involving two objects of the same class, FineART annotations disambiguate the objects $81.4\%$ of the time, compared with $44.0\%$ in ABC-130K. Such distinctions let a subtask specify which instance to manipulate, beyond naming the action and object class.

Together, object variation and instance-specific language provide supervision for selecting and manipulating objects throughout multi-stage tasks. \Cref{sec:experiments} evaluates how mid-training with these subtask annotations affects spatial grounding, instruction following, and transfer to a different embodiment.

\section{Experiments}
\label{sec:experiments}

We evaluate FineART as a mid-training dataset for a pre-trained generalist VLA
\cite{pi5}, mid-training for a relatively short run ($300$k steps) and then
asking three questions (\cref{sec:aloha-results,sec:yam}). Does dense subtask
training, together with knowledge insulation (KI), improve out-of-distribution
generalization and long-horizon instruction following? How much does
long-tail training data contribute to overall performance? And does FineART
mid-training reduce the data required to adapt to a new embodiment? Over the
course of our experiments, we execute $3{,}400$ rollouts.

\subsection{Policy and Mid-Training Setup}
\label{sec:policy-setup}

We start from $\pi_{0.5}$ publicly released via LeRobot \cite{pi5, lerobot}, whose
implementation does not support subtask training, knowledge insulation, or hierarchical
subtask-then-action inference. We extend the architecture to add all three and refer to the
resulting model as \textbf{FineART-VLA}: a System-2 subtask predictor and a System-1
continuous-action controller in one policy. It combines a $2$B-parameter Gemma backbone \cite{gemma2} with a
SigLIP vision encoder \cite{siglip} and a separate $300$M-parameter Gemma-style
action expert. Images and language form one bidirectional attention block; the state/action
suffix attends to itself and to that full block but not vice versa, and the subtask span being
generated is further restricted to its own earlier tokens, matching autoregressive generation at
inference. At step $t$ the policy observes $o_t = (I_t^{1:3}, q_t)$, three
of the rig's camera streams (bird's-eye and both wrists, resized to $224\times224$; the
worm's-eye view is recorded but not consumed) and the $14$-dim bimanual joint and gripper states. The policy is
conditioned on a language context $\ell$ that is either the task string $\ell^{\text{task}}$ or
a subtask $\ell^{\text{sub}}$.

Following \cite{pi5}, the policy emits three outputs per forward pass:
the next subtask in language, a FAST-tokenized action sequence $\tilde{a}$, and a continuous
chunk $A_t = [a_t, \dots, a_{t+H-1}] \in \mathbb{R}^{H\times 14}$ with $H=50$. Writing
$z_\theta \equiv z_\theta(o_t, \ell^{\text{sub}})$ for the backbone representation,
\begin{align}
\label{eq:policy}
&p_\theta\big(\ell^{\text{sub}}, \tilde{a}, A_t \mid o_t, \ell^{\text{task}}\big) = \nonumber\\
&\qquad p_\theta\big(\ell^{\text{sub}} \mid o_t, \ell^{\text{task}}\big)\;
p_\theta\big(\tilde{a} \mid o_t, \ell^{\text{sub}}\big)\;
p_\theta\big(A_t \mid z_\theta\big).
\end{align}
The continuous action head predicts a velocity $v_\theta$ from the noised chunk
$A_t^{\tau} = (1-\tau) A_t + \tau\epsilon$, trained by flow matching to minimize the
L2-norm between $v_\theta$ and the target field $\epsilon - A_t$, while the discrete and
text heads use next-token cross-entropy. The objective is
\begin{equation}
\label{eq:loss}
\mathcal{L}(\theta) = \lambda_{\text{flow}} \mathcal{L}_{\text{flow}}
+ \lambda_{\text{fast}} \mathcal{L}_{\text{fast}}
+ \lambda_{\text{text}} \mathcal{L}_{\text{text}},
\end{equation}
with $\lambda_{\text{flow}} = 10$, $\lambda_{\text{fast}} = 1$ and
$\lambda_{\text{text}} \in \{0,1\}$. Each action chunk in the dataset is represented in both the
continuous flow target and the discretized token sequence. Including explicit task-to-subtask predictions
($\lambda_{\text{text}} > 0$) has been shown to improve performance on long-horizon tasks, even compared to human oracle subtask planning \cite{pi5}. Setting
$\lambda_{\text{text}} = 0$ removes the subtask channel, conditioning the policy on $\ell^{\text{task}}$ directly.

Knowledge insulation (KI) \cite{KI} has been shown to mitigate the degradation of pre-trained
VLM knowledge that occurs when a randomly initialized flow-based action expert is attached to the
backbone. Insulating the backbone from $\mathcal{L}_{\text{flow}}$ restores
language-following, accelerates convergence by roughly $7.5\times$, and improves generalization to unseen objects and
environments. With KI
enabled, the action expert attends to stop-gradient copies of the backbone's keys and
values so that $\nabla_{\theta_{\text{VLM}}} \mathcal{L}_{\text{flow}} = 0$ and the
backbone is shaped only by the token-level losses.

When $\lambda_{\text{text}} =1 $, $30\%$ of each batch is a sample ($\ell^{\text{task}}$, $\ell^{\text{sub}}$,
$o_t$) for $\mathcal{L}_{\text{text}}$ and $70\%$ is a sample ($A_t$, $\tilde{a}$, $o_t$, $\ell^{\text{sub}}$) for $\mathcal{L}_{\text{flow}}$ and $\mathcal{L}_{\text{fast}}$.
All runs train for $300$k
steps at a global batch of $64$, about $10\%$ of one epoch over the $185.5$M frames, using
AdamW at $2.5\times10^{-5}$ decayed to $5\times10^{-6}$ in \texttt{bfloat16}. Mid-training draws
frames with a task-balanced sampler following \cite{pi5}: a task contributing $n_i$ frames is
sampled with probability proportional to $n_i^{\alpha}$, so $\alpha = 0$ weights every task
equally regardless of size (task-based sampling) and $\alpha = 1$ weights every frame equally
across the pooled dataset (raw frequency). We use $\alpha = 0.5$, a blend of the two.

\begin{table}[t]
\centering
\begin{threeparttable}
\footnotesize
\setlength{\tabcolsep}{2pt}
\renewcommand{\arraystretch}{1.15}
\begin{tabular}{@{}lcccccc@{\hspace{0.9em}}c@{}}
\toprule
& \multicolumn{2}{c}{\textbf{In-distribution}} & \multicolumn{2}{c}{\textbf{Partial ID}} & \multicolumn{2}{c}{\textbf{Out-of-distribution}} & \textbf{Overall} \\
\cmidrule(lr){2-3} \cmidrule(lr){4-5} \cmidrule(lr){6-7}
\textbf{Checkpoint} & Shelf & Towel & Vase & Pegboard & Saucer & Sort tools & Avg. \\
\midrule
\multicolumn{8}{@{}l}{\textit{(a) subtask training and knowledge insulation ($\alpha = 0.5$, full data)}} \\
Run 1.1 (subtask training + KI) & 91.0 / 72.0 & 72.0 / 32.0 & 75.0 / 16.0 & 55.0 / 0.0 & 73.0 / \textbf{20.0} & 69.5 / \textbf{28.0} & \textbf{72.6} / \textbf{28.0} \\
Run 1.2 (no subtask training + KI) & 97.0 / \textbf{96.0} & 50.0 / 16.0 & 74.0 / \textbf{24.0} & 72.0 / 0.0 & 33.0 / 4.0 & 76.5 / 16.0 & 67.1 / 26.0 \\
Run 1.3 (subtask training, no KI) & 77.0 / 64.0 & 86.0 / \textbf{44.0} & 66.0 / 16.0 & 49.0 / 0.0 & 69.0 / 12.0 & 58.5 / 12.0 & 67.6 / 24.7 \\
Run 1.4 (no subtask training, no KI) & 82.0 / 72.0 & 58.5 / 4.0 & 41.0 / 12.0 & 65.0 / 0.0 & 58.0 / 12.0 & 50.0 / 4.0 & 59.1 / 17.3 \\
\midrule
\multicolumn{8}{@{}l}{\textit{(b) Mid-training data ablations (w/ subtask training, no KI, $\alpha = 0.5$)}} \\
Run 1.3 (full data) & 77.0 / 64.0 & 86.0 / \textbf{44.0} & 66.0 / 16.0 & 49.0 / 0.0 & 69.0 / 12.0 & 58.5 / 12.0 & 67.6 / 24.7 \\
Run 2.1 (no long-tail\textsuperscript{$\dagger$}) & 88.0 / 68.0 & 29.5 / 0.0 & 67.0 / 8.0 & 58.0 / 0.0 & 56.0 / 0.0 & 67.0 / 8.0 & 60.9 / 14.0 \\
\bottomrule
\end{tabular}
\begin{tablenotes}[flushleft]\footnotesize
\item[]Each cell reports progress rate / success rate in percent over $25$ rollouts; the final column averages over all $150$ rollouts per checkpoint. Bold marks the best success rate in each column. \textsuperscript{$\dagger$}Run 2.1 restricts training to the two majority task-label categories in \cref{fig:dataset-overview}.
\end{tablenotes}
\caption{In-embodiment (ALOHA) results on the 6-task evaluation suite.}
\label{tab:aloha-main}
\end{threeparttable}
\end{table}

\begin{figure}[t]
\centering
\includegraphics[width=0.6\linewidth]{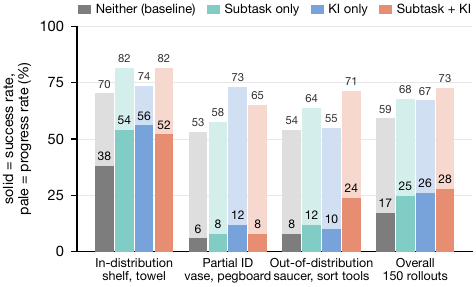}
\caption{Success rate of FineART-VLA by task regime (ID, partial ID, OOD) with subtask
supervision and knowledge insulation. ($n=25$ per task, $n=50$ per regime, $n=150$ overall). Subtask training and KI improve most significantly on OOD tasks.}
\label{fig:SS-KI}
\end{figure}

\subsection{In-Embodiment Results (ALOHA)}
\label{sec:aloha-results}

We train policy ablations via \cref{sec:policy-setup} on the full FineART dataset and evaluate the policies on the same embodiment used for collection (Stationary Trossen ALOHA). For all experiments, each checkpoint is evaluated with $25$ rollouts per task and is initialized with a prescribed set of $25$ object locations, meaning all checkpoints are evaluated under the same initial conditions.
We report
success rate (SR) and progress rate (PR, the average fraction of task stages completed).
At $25$ rollouts, per-task SR is
quantized to $4$ points with a standard error near $9$, and per-checkpoint averages carry
roughly $3$ to $4$, indicating large, consistent effects rather than single-task
differences.

\subsubsection{Does dense subtask training and knowledge insulation (KI) improve OOD generalization?}
We isolate the effect of subtask training (i.e., $\lambda_{\text{text}}=1$) and KI with a
$2\times2$ ablation that varies each independently: Run 1.1 (subtask training + KI), Run
1.2 (no subtask training + KI), Run 1.3 (subtask training, no KI), and Run 1.4 (no
subtask training, no KI), each evaluated on the same suite of six tasks.
The tasks are grouped in \cref{tab:aloha-main} by their relation to the tasks present in the FineART training set. Shelf and towel are
in-distribution, with literal subtask strings. Vase
and pegboard are partial, in that the insertion dynamics appear but the prompts differ. Saucer
and sort tools are out-of-distribution: the dataset contains no saucer and no type-routed
sorting of tools.

Comparing performance across in-distribution, partially in-distribution, and out-of-distribution tasks (\cref{fig:SS-KI}), subtask training and knowledge insulation each improve overall performance. Policies with both subtask training and KI show the highest OOD generalization capabilities, indicating that both subtask training and knowledge insulation are necessary to maintain general pre-trained task knowledge.

With in-distribution tasks, performance is dependent on the type of task (\cref{tab:aloha-main}).
Shelf placement is a saturated, single-step task, and no subtask training and KI achieves the highest performance ($96.0\%$ vs.\ $72.0\%$) indicating that subtask training may hinder task performance, while KI is important to maintain task understanding. Conversely, towel folding, which is a more ambiguous task, benefits from subtask training more than KI. KI and subtask training both improve performance compared to the baseline in Partial ID tasks, but benefit the most from KI alone; the comparatively worse performance with subtask training may indicate an over-sensitivity to language conditioning.

\subsubsection{How much does long-tail data contribute to overall performance?}
Run 2.1 (\cref{tab:aloha-main}b) removes the long-tail task categories defined in
\cref{fig:dataset-overview}c from mid-training, keeping the same number of training
steps; the retained set is only pick-and-place and sorting, $106$ of $151$ task labels
($71.0\%$ of valid frames), compared against Run 1.3 (full data, otherwise identical).
Overall success drops from $24.7\%$ to $14.0\%$. The simple shelf task is the exception,
holding its SR steady ($64.0\%$ to $68.0\%$), but every other evaluation task degrades
substantially when the long-tail is excluded, even though those tasks are nominally
covered by the retained categories -- indicating that long-tail diversity, not merely
category coverage, drives generalization.

\subsubsection{Does subtask training provide spatial grounding and long-horizon
instruction following?}
\begin{figure}[t]
\centering
\includegraphics[width=0.6\linewidth]{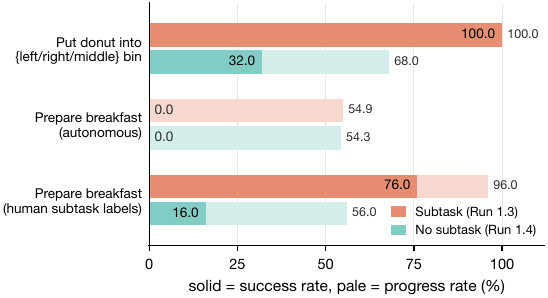}
\caption{Evaluation of subtask training on instruction following tasks (Run 1.3
against Run 1.4, no KI). The solid fill is success rate and the pale bar behind it
progress rate.}
\label{fig:subtask-heldout}
\end{figure}

In \cref{fig:subtask-heldout}, we further evaluate
Run 1.3 (subtask training, no KI) against Run 1.4 (no subtask training, no KI) -- KI
fixed off in both arms to isolate subtask training alone -- on two tasks designed to
test spatial grounding or long-horizon instruction following. The experimental procedure
is the same: $25$ rollouts per task with prescribed initial conditions.

\textbf{Spatial grounding.} On put donut into the \{left, right, middle\} bin, the target
is distinguished only by a spatial word in the instruction. Subtask training saturates the task with 100\% SR. In contrast, the no-subtask policy only achieves 32\% success and all 17 failures were due to placing the object in the incorrect bin. The results indicate that the dense subtask annotations in the FineART dataset are critical for spatial grounding and instruction following.

\textbf{Long-horizon instruction following.} Prepare breakfast is a multi-minute task
whose subtasks are present in the dataset, but not the full task sequence.
\Cref{fig:subtask-heldout} contrasts two inference conditions: autonomous execution, where the
policy proposes its own subtasks (Run 1.3, hierarchical subtask-then-action prediction
\cite{pi5}) or acts on the task label alone (Run 1.4), against human-provided subtask
guidance at inference, where a human oracle labels each subtask, advancing to the next
one only after the prior one is judged complete. Both autonomous policies fail
completely ($0.0\%$ success, at $54.9\%$ and $54.3\%$ progress respectively) since one
misstep is enough to stall the whole rollout. With human-provided subtask guidance, Run
1.3 reaches $76.0\%$ success against $16.0\%$ for Run 1.4 -- despite identical guidance,
so subtask training \textit{at training time}, not guidance at inference, is what drives
the gap.

\subsubsection{Observation: policies exhibit recovery and complex bimanual behaviors}
FineART-VLA reproduced notable behaviors that the dataset may contain in small proportion, but the task instruction does not explicitly call for (\cref{tab:emergent}).
\textit{Self-correction}: a sort tools policy redirected a trajectory already
heading for the wrong bin. \textit{Failure recovery}: a fallen cup was
re-grasped and the placement completed, instead of the rollout stalling on the
first error. \textit{Bimanual coordination}: on a flower in vase rollout, the left arm
steadied the vase so that the right could insert the flower. These behaviors indicate that policies are exhibiting closed-loop behavior and reacting to state feedback.

\begin{table}[t]
\centering
\begin{threeparttable}
\small
\setlength{\tabcolsep}{5pt}
\renewcommand{\arraystretch}{1.15}
\begin{tabular}{@{}l p{0.62\linewidth}@{}}
\toprule
\textbf{Task} & \textbf{Behavior} \\
\midrule
\multicolumn{2}{@{}l}{\textit{Run 1.1 (subtask training + KI)}} \\
Sort tools & Corrected in-flight trajectory toward wrong bin \\
\midrule
\multicolumn{2}{@{}l}{\textit{Run 1.2 (no subtask training + KI)}} \\
Sort tools & Recovery, and a handover between the arms \\
Sort tools & Handover between the arms \\
\midrule
\multicolumn{2}{@{}l}{\textit{Run 1.4 (no subtask training, no KI)}} \\
Fold towel & Completed two folds with the towel lifted \\
\midrule
\multicolumn{2}{@{}l}{\textit{Run 2.1 (no long-tail)}} \\
Cup on shelf & Re-grasped fallen cup and completed place \\
Flower in vase & Left arm steadied the vase so right could insert \\
\bottomrule
\end{tabular}
\begin{tablenotes}[flushleft]\footnotesize
\item \emph{Note: These are observations found via inspection and do not indicate the relative frequency of such behaviors.}
\end{tablenotes}
\caption{Notable behaviors observed in rollouts.}
\label{tab:emergent}
\end{threeparttable}
\end{table}

\subsection{Cross-Embodiment Transfer (YAM)}
\label{sec:yam}
\Cref{sec:aloha-results}
measures mid-trained policy performance on the in-distribution FineART embodiment.
In this section, we evaluate whether the mid-trained FineART-VLA shows similar improved performance on an \textit{out-of-distribution} embodiment. We fine-tune the
best-performing checkpoint from \cref{tab:aloha-main}a (Run 1.1, subtask training + KI,
$300$k steps) for an additional $5{,}000$ steps on YAM, using five of the six ALOHA
evaluation tasks -- shelf, towel, vase, pegboard, and saucer -- as the fine-tuning set.
Sort tools and prepare breakfast are held out entirely from YAM fine-tuning, to test
transfer to unseen tasks later in this section. We examine policies trained on $10$, $25$, $50$, and $250$ training episodes per task. Evaluations consist of $25$ rollouts each ($125$ rollouts per checkpoint). The embodiment gap is substantial: Stationary ALOHA arms are mounted on opposite sides of the table facing each other, whereas YAM arms face forward towards the back of the table. We compare against three alternatives fine-tuned
identically: the upstream $\pi_{0.5}$ initialization with no FineART mid-training, and two
checkpoints mid-trained on different datasets (ABC, and Ai2's MolmoAct2 YAM release). FineART is
the only one of the three datasets entirely on a different embodiment.

\begin{figure}[t]
\centering
\includegraphics[width=0.65\linewidth]{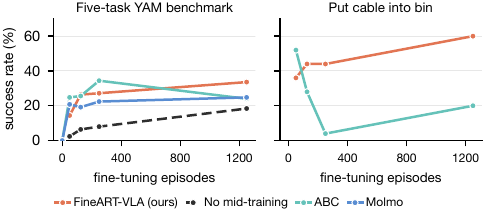}
\caption{FineART-VLA cross-embodiment YAM evaluation. Policies are mid-trained for 300k steps on the FineART, ABC, or MolmoAct2 datasets, then fine-tuned on YAM data for a fixed 5k steps at each data budget. Upstream $\pi_{0.5}$ open-source weights are not fine-tuned in mid-trained checkpoints. Since ABC and MolmoAct2 datasets contain YAM data, zero-shot (ZS) policies are evaluated for ABC and MolmoAct2 only.}
\label{fig:yam-scaling}
\end{figure}

\begin{table}[t]
\centering
\begin{threeparttable}
\small
\setlength{\tabcolsep}{5pt}
\renewcommand{\arraystretch}{1.15}
\begin{tabular}{@{}lccccc@{}}
\toprule
 & \multicolumn{5}{@{}c}{\textbf{Number of Fine-Tuning Trajectories}} \\
\textbf{Mid-Trained Checkpoint} & \textbf{Zero-shot} & \textbf{50} & \textbf{125} & \textbf{250} & \textbf{1250} \\
    & & \scriptsize(10/task) & \scriptsize(25/task) & \scriptsize(50/task) & \scriptsize(250/task) \\
    \midrule
\multicolumn{6}{@{}l}{\textit{(a) Five-task YAM benchmark}} \\
FineART-VLA (ours, ALOHA) & -- & 45.9 / 14.4 & 55.9 / \textbf{26.4} & 53.7 / 27.2 & 69.2 / \textbf{33.6} \\
None (upstream $\pi_{0.5}$ init) & -- & 25.0 / 2.4 & 33.4 / 6.4 & 36.1 / 8.0 & 50.8 / 18.4 \\
ABC (XDOF, YAM data) \textsuperscript{$\ddagger$} & 22.5 / 0.0 & 63.0 / \textbf{24.8} & 66.0 / 25.6 & 73.1 / \textbf{34.4} & 64.7 / 24.0 \\
MolmoAct2\textsuperscript{$\ddagger$} (Ai2, YAM data) & 18.2 / 0.0 & 61.1 / 20.8 & 66.7 / 19.2 & 62.4 / 22.4 & 68.1 / 24.8 \\
\midrule
\multicolumn{6}{@{}l}{\textit{(b) Put cable into bin, unseen target and unseen distractors}} \\
FineART-VLA (ours) & -- & 38.0 / 36.0 & 44.0 / \textbf{44.0} & 46.0 / \textbf{44.0} & 64.0 / \textbf{60.0} \\
ABC & -- & 52.0 / \textbf{52.0} & 28.0 / 28.0 & 4.0 / 4.0 & 22.0 / 20.0 \\
\bottomrule
\end{tabular}
\begin{tablenotes}[flushleft]\footnotesize
\item Cells are progress rate / success rate in percent. Block (a): average of $125$ rollouts across the first five tasks in \cref{tab:aloha-main}; block (b): single held-out task with $25$ rollouts. Bold
marks the best success rate per column. \textsuperscript{$\ddagger$}Zero-shot rates are reported only for ABC and MolmoAct2, whose mid-training datasets already include YAM data.
\end{tablenotes}
\caption{Cross-embodiment transfer to YAM, after a fixed $5$k-step fine-tune at each budget.}
\label{tab:yam}
\end{threeparttable}
\end{table}

\subsubsection{Does mid-training on FineART reduce embodiment-specific downstream data requirements?}
The FineART-VLA (orange) yields a $10\times$ reduction in downstream data compared to a checkpoint with no mid-training (black) in \cref{fig:yam-scaling}.
FineART-VLA reaches $26.4\%$ SR at $25$ fine-tuning episodes per task, compared to the $18.4\%$ SR at $250$ episodes per task when there is no mid-training.
Both checkpoints start from the same $\pi_{0.5}$ weights and differ only in whether the FineART mid-training occurred, so the gain comes from the mid-training stage rather than the underlying pre-training or model architecture.

\subsubsection{How does FineART-VLA performance on YAM (cross-embodiment) compare to policies from YAM open source datasets?}
We compare FineART-VLA to ABC and MolmoAct2, both of which include the YAM embodiment in their pre-training dataset.
On the five evaluation tasks in \cref{fig:yam-scaling}a, at the smallest budget FineART-VLA is the weakest
($14.4\%$ against $24.8\%$ for ABC and $20.8\%$ for MolmoAct2), which is consistent with the embodiment gap: ten episodes per task is not enough YAM data to re-target ALOHA kinematics.
FineART-VLA is the only checkpoint whose success rate increases monotonically to $33.6\%$ SR at $1250$ episodes.
MolmoAct2's SR plateaus in the low twenties despite a 25-fold increase in data (20.8\% to 24.8\%), while ABC peaks mid-sweep at 34.4\% before dropping to 24.0\%. Both baselines score 0.0\% success zero-shot, indicating task-embodiment fine-tuning is necessary regardless of mid-training dataset.

\subsubsection{Does FineART-VLA pick the right object among distractors on YAM?}
An additional task tests spatial grounding: put the cable into the bin, with a bread
roll and a donut as distractors, where the target and both distractors are absent from the YAM
fine-tuning data. FineART-VLA improves monotonically from $36.0\%$ to $60.0\%$ success as the
budget grows. ABC starts higher at ten episodes ($52.0\%$) but then drops sharply as the fine-tuning set grows, suggesting the representation is overwritten, not refined, by
fine-tuning.

\begin{figure}[t]
\centering
\includegraphics[width=0.6\linewidth]{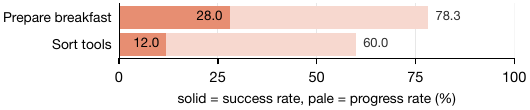}
\caption{FineART-VLA transfer to tasks absent from the YAM five-task fine-tuning set; 300k ALOHA mid-training steps + 5k fine-tuning steps on YAM.}
\label{fig:yam-unseen}
\end{figure}

\subsubsection{Does fine-tuning on non-relevant YAM tasks enable transfer to unseen long-horizon tasks?}
The most striking result is on tasks that appear nowhere in the YAM fine-tuning data. After the
same $5$k-step fine-tune on the five basic tasks (\cref{fig:yam-scaling}a), we evaluate FineART-VLA on prepare breakfast, which requires
placing a plate on the tray and then arranging two donuts and a slice of bread on it. As shown in \cref{fig:yam-unseen}, the
policy reaches $28.0\%$ SR and $78.3\%$ PR with autonomous hierarchical execution, executing the full multi-stage sequence
on a robot it had seen only through five unrelated tasks. It also emits the \textit{Adjust}
subtask to realign the tray, a behavior annotated in the FineART ALOHA data and never performed in the
YAM fine-tuning set. Sort tools, also unseen on YAM, reaches $12.0\%$ SR and $60.0\%$ PR.
These results indicate that subtask-level task structure transfers across the embodiment gap, and the fine-tuning trajectories supply the kinematics for embodiment-specific motor patterns.

\section{Conclusion}
\label{sec:conclusion}

To our knowledge, FineART is the largest open-source subtask-annotated manipulation dataset: $533{,}913$ labels covering every one of its $40{,}543$ episodes and $151$ tasks, spanning hundreds of distinct object classes plus a long tail of contact-rich skills beyond ordinary pick-and-place tasks (\cref{sec:FineART}).

Experiments with our FineART-VLA policy indicate that the FineART dataset (\cref{sec:experiments}) enables meaningful performance gains on in-distribution and out-of-distribution tasks and across embodiments.
Inclusion of long-tail tasks enables higher performance across partially in-distribution and out-of-distribution evaluation tasks, indicating object variety alone is not sufficient for task diversity. Subtask training enables FineART-VLA to parse instructions, which a flat policy cannot, and to execute long-horizon sequential tasks. In cross-embodiment experiments with minimal target embodiment fine-tuning (Stationary table-top ALOHA to Mobile YAM), our FineART-VLA policy completes long-horizon tasks not present in the target embodiment fine-tuning set. These results indicate that FineART-VLA learned subtask-level planning during ALOHA mid-training, and the target embodiment fine-tuning trains the policy on the target robot's kinematics without forgetting the sequential task planning.

\paragraph{Limitations and Future Work.} Our dataset focuses on a single robot platform, and our policy is only trained for about $10\%$ of an epoch. Training for longer and testing more base models would likely show a larger impact of dense subtask annotations. Given the impressive performance gains from our subtask-conditioned FineART-VLA policy, we are excited to see progress on subtask-conditioned world models \cite{ctrlworld}, reward models \cite{sarm}, and advantage-conditioned RL \cite{recap}, which enables learning from rejected rollouts instead of discarding them. VLM-automated labeling shows promise for reducing subtask annotation burden as the dataset grows \cite{nils}. Finally, while fine-tuning on five non-relevant YAM tasks enabled transfer to unseen, long-horizon YAM tasks, it is unclear which types of fine-tuning tasks drove cross-embodiment transfer. A task, annotation, and embodiment sweep \cite{oxe, crossformer} would help isolate the effects of mid-training and fine-tuning dataset diversity. By releasing the FineART dataset, model, and training code, we present a canvas for future exploration in our field.

\section*{Acknowledgments}
We thank our robot operators for data collection, annotation, evaluations, and logistics, and the many colleagues across engineering and program leadership who supported this work. See \cref{app:ack} for full acknowledgments.

\bibliographystyle{abbrvnat}
\bibliography{references}

\clearpage
\appendix

\section*{Appendix}

\section{FineART-VLA: Architecture and Mid-Training Details}
\label{app:architecture}

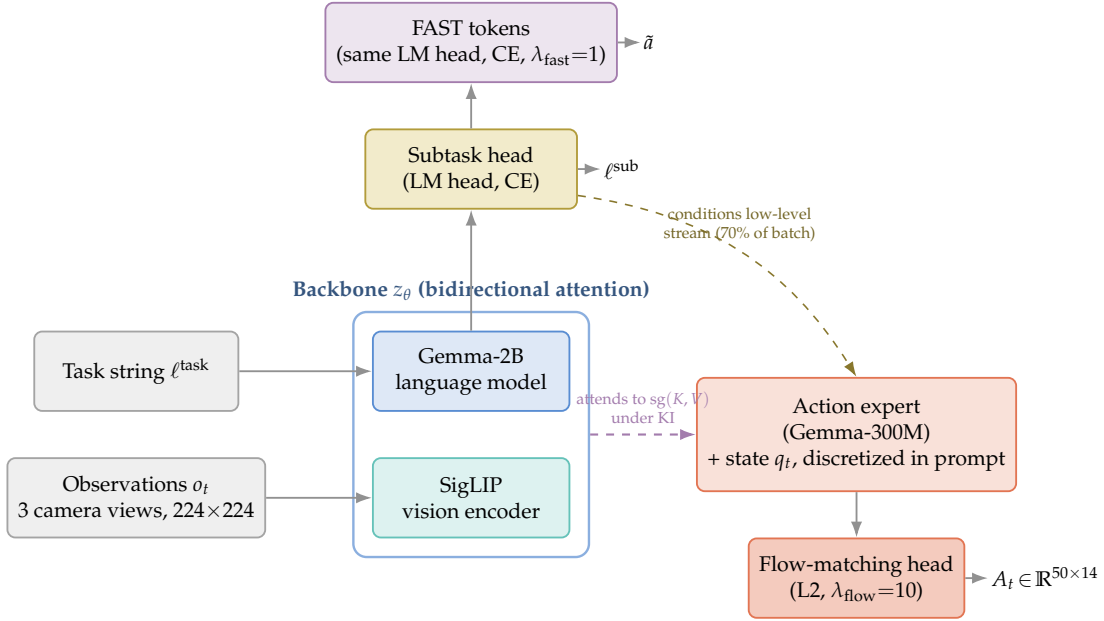
\begin{figure}[H]
\centering
\begin{tikzpicture}[
  font=\scriptsize,
  node distance=6mm and 14mm,
  every node/.style={align=center},
  block/.style={draw, rounded corners=3pt, line width=0.7pt, minimum height=1.05cm, inner sep=4pt},
  obsblock/.style={block, fill=scaleGray!12, draw=scaleGray!70, minimum width=2.7cm},
  visionblock/.style={block, fill=scaleTeal!22, draw=scaleTeal, minimum width=2.6cm},
  lmblock/.style={block, fill=scaleBlue!20, draw=scaleBlue, minimum width=2.6cm},
  backbonebox/.style={draw=scaleBlue!70, rounded corners=5pt, line width=0.9pt, inner sep=7pt},
  textheadblock/.style={block, fill=scaleGold!28, draw=scaleGold!85!black, minimum width=2.8cm},
  expertblock/.style={block, fill=scaleTerracotta!22, draw=scaleTerracotta, minimum width=2.8cm, minimum height=1.5cm},
  fastheadblock/.style={block, fill=scalePurple!20, draw=scalePurple, minimum width=2.7cm},
  flowheadblock/.style={block, fill=scaleTerracotta!38, draw=scaleTerracotta, minimum width=2.7cm},
  lbl/.style={draw=none, fill=none, inner sep=1pt},
  arr/.style={-{Latex[length=2.1mm,width=1.5mm]}, line width=0.7pt, draw=scaleGray!85},
  kiarr/.style={-{Latex[length=2.1mm,width=1.5mm]}, line width=0.8pt, draw=scalePurple, dashed},
  fbarr/.style={-{Latex[length=2mm,width=1.4mm]}, line width=0.6pt, draw=scaleGold!65!black, dashed},
]

\node[obsblock] (obs) {Observations $o_t$\\{\scriptsize 3 camera views, $224{\times}224$}};
\node[obsblock, above=of obs] (task) {Task string $\ell^{\text{task}}$};

\node[visionblock, right=of obs] (vision) {SigLIP\\vision encoder};
\node[lmblock, above=6mm of vision] (lm) {Gemma-2B\\language model};
\node[backbonebox, fit=(vision)(lm)] (backbone) {};
\node[lbl, above=1mm of backbone, font=\scriptsize\bfseries, text=scaleBlue!60!black] (backbonelabel) {Backbone $z_\theta$ (bidirectional attention)};

\node[textheadblock, above=16mm of lm] (subtask) {Subtask head\\(LM head, CE)};
\node[lbl, right=3mm of subtask] (lsub) {$\ell^{\text{sub}}$};

\node[fastheadblock, above=of subtask] (fast) {FAST tokens\\(same LM head, CE, $\lambda_{\text{fast}}{=}1$)};
\node[lbl, right=3mm of fast] (atilde) {$\tilde a$};

\node[expertblock, right=of backbone] (expert) {Action expert\\(Gemma-300M)\\{\scriptsize + state $q_t$, discretized in prompt}};

\node[flowheadblock, below=of expert] (flow) {Flow-matching head\\(L2, $\lambda_{\text{flow}}{=}10$)};

\node[lbl, right=3mm of flow] (achunk) {$A_t\!\in\!\mathbb{R}^{50\times14}$};

\draw[arr] (obs.east) -- (vision.west);
\draw[arr] (task.east) -- (lm.west);
\draw[arr] (lm.north) -- (subtask.south);
\draw[arr] (subtask.east) -- (lsub.west);
\draw[arr] (subtask.north) -- (fast.south);
\draw[arr] (fast.east) -- (atilde.west);

\draw[fbarr] ([yshift=-3.5mm]subtask.east) to[bend left=25] node[midway, above, font=\tiny, align=center, text=scaleGold!55!black]{conditions low-level\\stream ($70\%$ of batch)} (expert.north);

\draw[kiarr] (backbone.east) -- node[above, font=\tiny, text=scalePurple, align=center]{attends to $\mathrm{sg}(K,V)$\\under KI} (expert.west);

\draw[arr] (expert.south) -- (flow.north);
\draw[arr] (flow.east) -- (achunk.west);

\end{tikzpicture}
\caption{FineART-VLA architecture. The backbone $z_\theta$ is a
$2$B-parameter Gemma language model with a SigLIP vision encoder, attending
bidirectionally over the camera views and the task string. Its LM head
decodes the subtask $\ell^{\text{sub}}$ and, right after it, the
FAST-tokenized action sequence $\tilde a$. The separate $300$M-parameter
action expert reads out the continuous chunk $A_t$ through its own
flow-matching loss. Under knowledge insulation, the action expert attends to
stop-gradient copies of the backbone's keys and values.}
\label{fig:architecture}
\end{figure}

\Cref{fig:architecture} gives the full computational path behind
\cref{sec:policy-setup}'s architecture, \cref{fig:attention-mask} gives the
attention masks behind \cref{eq:policy}'s two training shapes, and
\cref{tab:training-config} gives the mid-training configuration behind
all of our subtask/KI runs in \cref{tab:aloha-main}(a).

The LM head decodes the subtask on $30\%$ of samples ($\lambda_{\text{text}}{=}1$), and the action expert conditions on it for the remaining $70\%$
(\cref{eq:policy}). The action expert always conditions on the $14$-dim joint state $q_t$, discretized into $256$ bins and embedded as text in the same prompt
rather than passed through the backbone. The FAST and the
flow-matching output never attend to each other. With knowledge
insulation on, the flow loss's gradient never reaches $\theta_{\text{VLM}}$
while the FAST loss -- computed entirely by the backbone's own LM head -- is
unaffected either way (\cref{app:ki}). Because a sample either generates the
subtask ($30\%$) or predicts the FAST and action outputs conditioned on it
($70\%$), these two cases use two different attention patterns rather than
one shared sequence, shown side by side in \cref{fig:attention-mask}
(\cref{app:ki}).

\begin{table}[H]
\centering
\begin{threeparttable}
\footnotesize
\setlength{\tabcolsep}{4pt}
\renewcommand{\arraystretch}{1.12}
\begin{tabular}{@{}ll@{}}
\toprule
\textbf{Setting} & \textbf{Value} \\
\midrule
\multicolumn{2}{@{}l}{\textit{(a) Mid-training data}} \\
Dataset & FineART, task-label-repaired release \\
Episodes / frames & $40{,}543$ / $185{,}534{,}500$ \\
Tasks & $151$ (dense task index) \\
Cameras & bird's-eye, left wrist, right wrist ($360{\times}640$, resized $224{\times}224$) \\
State / action & $14$-dim bimanual joint + gripper, per frame \\
Sampling & task-balanced, $\alpha = 0.5$, idle chunks stripped ($\approx 3.1\%$ of frames) \\
\midrule
\multicolumn{2}{@{}l}{\textit{(b) Optimization}} \\
Optimizer & AdamW ($\beta_1{=}0.9$, $\beta_2{=}0.95$, $\epsilon{=}10^{-8}$) \\
Learning rate & $2.5\times10^{-5}$, cosine decay to $5\times10^{-6}$ \\
Warmup & $2{,}000$ steps \\
Weight decay & $10^{-10}$ \\
Gradient clip & $1.0$ (global norm) \\
Text CE $z$-loss & weight $10^{-4}$ (large-vocab logit regularizer) \\
Precision & native \texttt{bfloat16} (no autocast), gradient checkpointing \\
\midrule
\multicolumn{2}{@{}l}{\textit{(c) Compute}} \\
Hardware & $1$ node, $8\times$H100 per run \\
Global batch & $64$ ($8$ per device $\times$ $8$ devices) \\
Steps & $300{,}000$ ($\approx 10\%$ of one epoch) \\
Wall-clock & $\approx 1$ week per run (four runs trained in parallel) \\
Checkpointing & every $2{,}500$ steps ($120$ checkpoints saved) \\
\bottomrule
\end{tabular}
\begin{tablenotes}[flushleft]\footnotesize
\item[]All four subtask/KI ablation runs in \cref{tab:aloha-main}(a) share this configuration and differ only in the recipe (subtask training on/off) and whether knowledge insulation is enabled. Per-component learning-rate multipliers for the LM head, backbone, and action expert are all $1.0$ (no differential scaling).
\end{tablenotes}
\caption{Mid-training configuration for FineART-VLA.}
\label{tab:training-config}
\end{threeparttable}
\end{table}

\subsection{Task-Balanced Sampling}
\label{app:sampling}

FineART's $151$ tasks range from a few dozen episodes to tens of thousands, so
sampling frames uniformly from the pooled dataset would spend most of
training on whichever tasks happen to be over-represented, starving all others. We use the temperature-sampling
scheme from \cite{pi5} behind the ``task-balanced, $\alpha=0.5$'' row of
\cref{tab:training-config}(a): task $i$, with $n_i$ valid (non-idle) frames,
is drawn with probability

\begin{equation}
\label{eq:task-sampling}
p(\text{task}=i) = \frac{n_i^{\alpha}}{\sum_j n_j^{\alpha}},
\end{equation}

and once a task is chosen, the frame itself is drawn uniformly from that
task's $n_i$ valid frames. $\alpha$ interpolates between two extremes. At
$\alpha=0$, \cref{eq:task-sampling} gives $p(\text{task}=i) = 1/151$ for every
task regardless of size, so a task with a handful of episodes gets exactly as
many gradient steps as one an order of magnitude larger. At $\alpha=1$, it
reduces to $n_i / \sum_j n_j$, i.e., sampling frames uniformly from the
pooled dataset, so the largest tasks dominate in direct proportion to their
size. We use $\alpha=0.5$ for every run in this paper: a task twice as large
as another is still sampled more often, but only by a factor of
$\sqrt{2}$ rather than $2$.

We do not implement \cref{eq:task-sampling} as a single per-frame weight
vector handed to a weighted sampler since building and drawing from a
categorical distribution over FineART's $185.5$M frames is beyond what
\texttt{torch.multinomial}'s categorical cap ($2^{24}$ entries) supports.
We sample hierarchically: draw a task from \cref{eq:task-sampling} by
an inverse-CDF lookup on the $151$-entry task distribution, then draw a frame
uniformly from a precomputed per-task index of valid frames, costing two
small lookups per sample instead of one lookup into a $185.5$M-entry table.

\subsection{Knowledge Insulation}
\label{app:ki}

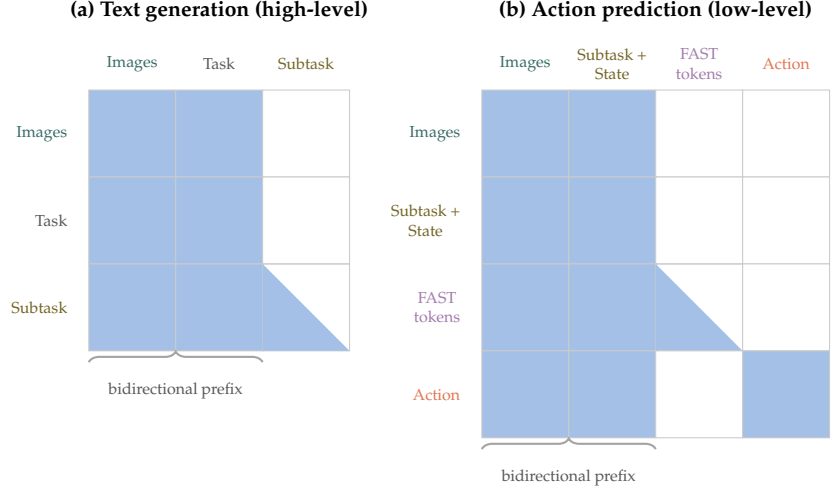
\begin{figure}[H]
\centering
\begin{tikzpicture}[
  font=\scriptsize,
  every node/.style={align=center},
]

\node[font=\scriptsize\bfseries] at (1.725,1.05) {(a) Text generation (high-level)};

\node[font=\tiny, text=scaleTeal!55!black] at (0.575,0.35) {Images};
\node[font=\tiny, text=scaleGray!70!black] at (1.725,0.35) {Task};
\node[font=\tiny, text=scaleGold!55!black] at (2.875,0.35) {Subtask};

\node[font=\tiny, text=scaleTeal!55!black, anchor=east] at (-0.15,-0.575) {Images};
\node[font=\tiny, text=scaleGray!70!black, anchor=east] at (-0.15,-1.725) {Task};
\node[font=\tiny, text=scaleGold!55!black, anchor=east] at (-0.15,-2.875) {Subtask};

\attendcell{0}{0}      \attendcell{1.15}{0}      \maskcell{2.3}{0}
\attendcell{0}{-1.15}  \attendcell{1.15}{-1.15}  \maskcell{2.3}{-1.15}
\attendcell{0}{-2.3}   \attendcell{1.15}{-2.3}   \causalcell{2.3}{-2.3}

\draw[decorate, decoration={brace, amplitude=4pt}, thick, scaleGray!70]
  (0,-3.6) -- (2.3,-3.6)
  node[midway, yshift=-0.38cm, font=\tiny, text=scaleGray!70!black]{bidirectional prefix};

\begin{scope}[xshift=5.2cm]
\node[font=\scriptsize\bfseries] at (2.3,1.05) {(b) Action prediction (low-level)};

\node[font=\tiny, text=scaleTeal!55!black] at (0.575,0.35) {Images};
\node[font=\tiny, text=scaleGold!55!black] at (1.725,0.35) {Subtask +\\State};
\node[font=\tiny, text=scalePurple] at (2.875,0.35) {FAST\\tokens};
\node[font=\tiny, text=scaleTerracotta] at (4.025,0.35) {Action};

\node[font=\tiny, text=scaleTeal!55!black, anchor=east] at (-0.15,-0.575) {Images};
\node[font=\tiny, text=scaleGold!55!black, anchor=east] at (-0.15,-1.725) {Subtask +\\State};
\node[font=\tiny, text=scalePurple, anchor=east] at (-0.15,-2.875) {FAST\\tokens};
\node[font=\tiny, text=scaleTerracotta, anchor=east] at (-0.15,-4.025) {Action};

\attendcell{0}{0}      \attendcell{1.15}{0}      \maskcell{2.3}{0}      \maskcell{3.45}{0}
\attendcell{0}{-1.15}  \attendcell{1.15}{-1.15}  \maskcell{2.3}{-1.15}  \maskcell{3.45}{-1.15}
\attendcell{0}{-2.3}   \attendcell{1.15}{-2.3}   \causalcell{2.3}{-2.3} \maskcell{3.45}{-2.3}
\attendcell{0}{-3.45}  \attendcell{1.15}{-3.45}  \maskcell{2.3}{-3.45}  \attendcell{3.45}{-3.45}

\draw[decorate, decoration={brace, amplitude=4pt}, thick, scaleGray!70]
  (0,-4.75) -- (2.3,-4.75)
  node[midway, yshift=-0.38cm, font=\tiny, text=scaleGray!70!black]{bidirectional prefix};
\end{scope}

\end{tikzpicture}
\caption{Block-causal attention pattern. Colored cells can attend
to each other, white cells are masked, and the diagonal split means
attention is causal within that block. \textbf{(a)} Generating the subtask
($30\%$ of samples). Images and the task string form the bidirectional
prefix, and the subtask is generated one token at a time, so each subtask
token can only see earlier subtask tokens -- the autoregressive restriction
described in \cref{sec:policy-setup}. \textbf{(b)} Predicting the action
($70\%$ of samples, or every sample for the no-subtask arm). Here the
subtask and discretized state are given as input rather than generated, so
they join the bidirectional prefix instead. The FAST-tokenized sequence is
still generated token by token, so it stays causal like the subtask span in
\textbf{(a)}. The continuous action tokens have no such restriction: flow
matching denoises the whole chunk at once, so they can all attend freely to
each other.}
\label{fig:attention-mask}
\end{figure}

\begin{figure}[H]
\centering
\begin{tikzpicture}[
  font=\scriptsize,
  node distance=6mm and 12mm,
  every node/.style={align=center},
  block/.style={draw, rounded corners=3pt, line width=0.7pt, minimum height=0.9cm, inner sep=4pt},
  vlmblock/.style={block, fill=scaleBlue!20, draw=scaleBlue, minimum width=2.7cm},
  actblock/.style={block, fill=scaleTerracotta!22, draw=scaleTerracotta, minimum width=2.7cm},
  sgblock/.style={block, fill=scalePurple!15, draw=scalePurple, dashed, minimum width=2.7cm},
  lbl/.style={draw=none, fill=none, inner sep=1pt},
  arr/.style={-{Latex[length=2.1mm,width=1.5mm]}, line width=0.7pt, draw=scaleGray!85},
  kiarr/.style={-{Latex[length=2.1mm,width=1.5mm]}, line width=0.8pt, draw=scalePurple, dashed},
  blockedarr/.style={-{Latex[length=2mm,width=1.4mm]}, line width=0.7pt, draw=red!70!black, dashed},
]

\node[lbl] (hvlm0) {$h_{\text{vlm}}^{(l-1)}$};
\node[vlmblock, right=of hvlm0] (qkvvlm) {$Q_{\text{vlm}}, K_{\text{vlm}}, V_{\text{vlm}}$};
\node[vlmblock, right=of qkvvlm] (attvlm) {$\mathrm{Att}_{\text{vlm}}^{(l)}$};
\node[lbl, right=of attvlm] (hvlm1) {$h_{\text{vlm}}^{(l)}$};

\node[lbl, below=18mm of hvlm0] (hact0) {$h_{\text{act}}^{(l-1)}$};
\node[actblock, right=of hact0] (qkvact) {$Q_{\text{act}}, K_{\text{act}}, V_{\text{act}}$};
\node[actblock, right=of qkvact] (attact) {$\mathrm{Att}_{\text{act}}^{(l)}$};
\node[lbl, right=of attact] (hact1) {$h_{\text{act}}^{(l)}$};

\node[sgblock, below=6mm of qkvvlm] (sg) {$\mathrm{sg}(K_{\text{vlm}}, V_{\text{vlm}})$};

\draw[arr] (hvlm0) -- (qkvvlm);
\draw[arr] (qkvvlm) -- (attvlm);
\draw[arr] (attvlm) -- (hvlm1);

\draw[arr] (hact0) -- (qkvact);
\draw[arr] (qkvact) -- (attact);
\draw[arr] (attact) -- (hact1);

\draw[arr] (qkvvlm) -- (sg);
\draw[kiarr] (sg) -- (attact);

\draw[blockedarr] (attact.north) to[bend right=30] (sg.south east);
\node[font=\tiny, text=red!70!black, xshift=1pt, yshift=1pt] at (sg.south east) {$\times$};
\node[font=\tiny, text=red!70!black, above=1mm of sg, xshift=10mm] {blocked};

\end{tikzpicture}
\caption{Knowledge insulation at layer $l$. The VLM stream's attention
$\mathrm{Att}_{\text{vlm}}^{(l)}$ (top) is untouched by KI. The action
stream's attention $\mathrm{Att}_{\text{act}}^{(l)}$ (bottom) reads the VLM
stream's keys and values only through the stop-gradient copy
$\mathrm{sg}(K_{\text{vlm}}, V_{\text{vlm}})$ (purple, dashed), which blocks
$\nabla_{\theta_{\text{VLM}}}\mathcal{L}_{\text{flow}}$ at the red $\times$
(\cref{eq:ki-grad}).}
\label{fig:ki}
\end{figure}
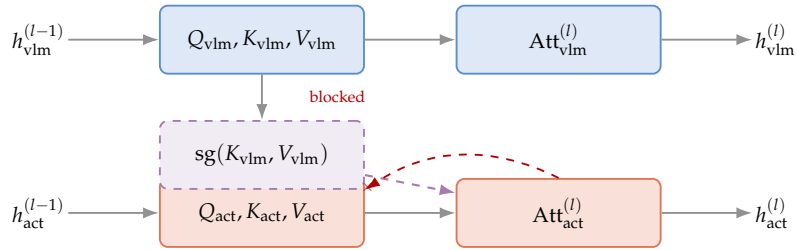

\Cref{sec:policy-setup} asserts that knowledge insulation makes
$\nabla_{\theta_{\text{VLM}}}\mathcal{L}_{\text{flow}} = 0$. Here we show why.
At every layer $l = 1,\dots,L$ of the joint attention, drop the head
dimension and write the VLM stream's own query/key/value as
$Q_{\text{vlm}}^{(l)}, K_{\text{vlm}}^{(l)}, V_{\text{vlm}}^{(l)}$
(projections of the backbone hidden state $h_{\text{vlm}}^{(l-1)}$) and the
action expert's as $Q_{\text{act}}^{(l)}, K_{\text{act}}^{(l)},
V_{\text{act}}^{(l)}$ (projections of $h_{\text{act}}^{(l-1)}$, its own
$300$M-parameter Gemma tower). Because the prefix never attends into the
suffix (\cref{fig:attention-mask}), the VLM stream's attention is ordinary
self-attention,
\begin{equation}
\mathrm{Att}_{\text{vlm}}^{(l)} = \mathrm{softmax}\!\left(
\frac{Q_{\text{vlm}}^{(l)}\,K_{\text{vlm}}^{(l)\top}}{\sqrt{d}} + M_{\text{vlm}}
\right) V_{\text{vlm}}^{(l)},
\end{equation}
and its final layer's output realizes the backbone representation $z_\theta$
of \cref{eq:policy}. The action stream's attention additionally reads the VLM
stream's keys and values (it is conditioned on the backbone, not vice versa):
\begin{equation}
\label{eq:ki-attn}
\mathrm{Att}_{\text{act}}^{(l)} = \mathrm{softmax}\!\left(
\frac{Q_{\text{act}}^{(l)}\,[\kappa^{(l)}; K_{\text{act}}^{(l)}]^\top}{\sqrt{d}}
+ M_{\text{act}}\right) [\nu^{(l)}; V_{\text{act}}^{(l)}],
\end{equation}
where $[\cdot\,;\,\cdot]$ concatenates along the key/value sequence axis and
$M_{\text{vlm}}, M_{\text{act}}$ are the additive masks realizing
\cref{fig:attention-mask}(b). \textbf{Without KI}, $\kappa^{(l)} =
K_{\text{vlm}}^{(l)}$ and $\nu^{(l)} = V_{\text{vlm}}^{(l)}$ directly.
\textbf{With KI}, $\kappa^{(l)} = \mathrm{sg}\big(K_{\text{vlm}}^{(l)}\big)$
and $\nu^{(l)} = \mathrm{sg}\big(V_{\text{vlm}}^{(l)}\big)$, where
$\mathrm{sg}(\cdot)$ is the stop-gradient operator: the identity in the
forward pass ($\mathrm{sg}(x) = x$) but a constant under differentiation
($\partial\,\mathrm{sg}(x)/\partial x \equiv 0$). \Cref{fig:ki} draws this
split for a single layer.

Because $\mathrm{sg}(\cdot)$ is the identity in the forward direction,
\cref{eq:ki-attn} takes the exact same numerical value whether or not KI is
enabled. KI is a purely backward-pass ablation, changing which parameters
receive gradient and not what the model computes. Let $\theta_{\text{VLM}}$ denote every backbone parameter
feeding $K_{\text{vlm}}^{(l)}, V_{\text{vlm}}^{(l)}$ at any layer, and note
that $\mathcal{L}_{\text{flow}}$ (\cref{eq:loss}) is read out of the action
stream's outputs only, $\mathcal{L}_{\text{flow}} = g\big(
\mathrm{Att}_{\text{act}}^{(1)}, \dots, \mathrm{Att}_{\text{act}}^{(L)}\big)$,
never touching $\mathrm{Att}_{\text{vlm}}^{(l)}$ directly. With KI, for every
layer $l$,
\begin{equation}
\label{eq:ki-grad}
\frac{\partial\,\mathrm{Att}_{\text{act}}^{(l)}}{\partial\theta_{\text{VLM}}} =
\frac{\partial\,\mathrm{Att}_{\text{act}}^{(l)}}{\partial\,\mathrm{sg}(K_{\text{vlm}}^{(l)})}
\underbrace{\frac{\partial\,\mathrm{sg}(K_{\text{vlm}}^{(l)})}{\partial\theta_{\text{VLM}}}}_{\equiv\,0}
\;+\;
\frac{\partial\,\mathrm{Att}_{\text{act}}^{(l)}}{\partial\,\mathrm{sg}(V_{\text{vlm}}^{(l)})}
\underbrace{\frac{\partial\,\mathrm{sg}(V_{\text{vlm}}^{(l)})}{\partial\theta_{\text{VLM}}}}_{\equiv\,0}
\;=\; 0,
\end{equation}
so by the chain rule $\nabla_{\theta_{\text{VLM}}}\mathcal{L}_{\text{flow}} =
\sum_{l=1}^{L} \frac{\partial \mathcal{L}_{\text{flow}}}{\partial\,
\mathrm{Att}_{\text{act}}^{(l)}} \cdot \frac{\partial\,
\mathrm{Att}_{\text{act}}^{(l)}}{\partial\theta_{\text{VLM}}} = 0$. This split
is applied uniformly at every one of the $L$ backbone layers (once per layer
of the fused forward), so no VLM parameter at any depth -- not just a given
layer's own $K$/$V$ projection weights, but every earlier embedding and layer
that produced $h_{\text{vlm}}^{(l-1)}$ -- can reach
$\mathcal{L}_{\text{flow}}$ through this path either. $\mathrm{sg}(\cdot)$
severs the graph at that node regardless of what feeds into it upstream.

The FAST-tokenized output (\cref{fig:architecture}) is computed from the
\textit{VLM} stream, not the action stream. FAST tokens are appended to the prefix's own text
tokens before layer $1$ (\cref{fig:attention-mask}(b): ``FAST tokens''
shares the bidirectional-prefix side with ``Subtask + State'') and read out
through $\mathrm{Att}_{\text{vlm}}^{(L)}$ using the same PaliGemma LM head
that predicts the subtask. Both
$\mathcal{L}_{\text{text}}$ and $\mathcal{L}_{\text{fast}}$ are sliced from
the same VLM-stream output tensor, which \cref{eq:ki-attn} never modifies.
So
\begin{equation}
\nabla_{\theta_{\text{VLM}}}\mathcal{L}_{\text{text}} \neq 0, \qquad
\nabla_{\theta_{\text{VLM}}}\mathcal{L}_{\text{fast}} \neq 0,
\end{equation}
by the same argument in reverse: neither loss's computational graph passes
through an $\mathrm{sg}(\cdot)$ node, so the backbone continues to receive
full gradient from both. KI insulates the backbone from the \textit{continuous}
action loss only, not from the discrete one -- consistent with ``the
backbone is shaped only by the token-level losses'' (\cref{sec:policy-setup}).

In no panel does the prefix attend back into a generated or predicted
suffix, and the FAST and continuous-action outputs are mutually invisible
so neither loss leaks into the other. This masking pattern is fixed by the
recipe regardless of knowledge insulation.

\subsection{Optimization Objective}
\label{app:objective}

\Cref{eq:loss} gives the three loss terms. In practice all of them are computed
on every action chunk regardless of $\lambda_{\text{text}}$ since the same
chunk is flow-matched and FAST-tokenized at once. We also put a $z$-loss on the
text logits, weight $10^{-4}$: without it, a large-vocabulary head like
PaliGemma's $257$K tokens can let its log-partition function drift while
cross-entropy itself persists. The recipe stack can drop the subtask
context at random during training, so the policy learns to fall back on the
task string alone. We leave that dropout at $0$ for every run in this
paper.

Mid-training required $300$K steps at a global batch of $64$ -- about $10\%$
of one epoch over FineART's $185.5$M frames -- roughly a week of wall-clock
time on a single $8\times$H100 node. We ran the
four subtask/KI ablations on separate nodes at the same time rather than one
after another, so all four checkpoints in \cref{tab:aloha-main}(a) were ready
within that same week.

\subsection{Inference}
\label{app:inference}

\begin{figure}[t]
\centering
\begin{tikzpicture}[
  font=\scriptsize,
  every node/.style={align=center},
  block/.style={draw, rounded corners=3pt, line width=0.7pt, minimum height=1.05cm, inner sep=4pt, minimum width=3.3cm},
  taskbox/.style={block, fill=scaleGray!12, draw=scaleGray!70},
  lmbox/.style={block, fill=scaleGold!28, draw=scaleGold!85!black},
  opbox/.style={block, fill=scaleSky!25, draw=scaleSky!70!black},
  expertbox/.style={block, fill=scaleTerracotta!22, draw=scaleTerracotta, minimum height=1.3cm},
  hdr/.style={font=\scriptsize\bfseries, draw=none, fill=none},
  note/.style={font=\tiny, draw=none, fill=none, text width=3.5cm, align=center},
  arr/.style={-{Latex[length=2.1mm,width=1.5mm]}, line width=0.7pt, draw=scaleGray!85},
  goldarr/.style={-{Latex[length=2.1mm,width=1.5mm]}, line width=0.7pt, draw=scaleGold!70!black},
]

\node[taskbox] (task-a) at (0,0) {Task $\ell^{\text{task}}$};
\node[expertbox] (expert-a) at (0,3.6) {Action expert\\+ flow head\\$\to A_t$};
\node[hdr] at (0,4.6) {(a) Flat};
\draw[arr] (task-a) -- (expert-a);
\node[note] at (0,-1.3) {Re-conditions on raw task string for every chunk without LM call};

\node[taskbox] (task-b) at (5.4,0) {Task $\ell^{\text{task}}$};
\node[lmbox] (lm-b) at (5.4,1.8) {LM head\\generates $\ell^{\text{sub}}$};
\node[expertbox] (expert-b) at (5.4,3.6) {Action expert\\+ flow head\\$\to A_t$};
\node[hdr] at (5.4,4.6) {(b) Hierarchical};
\draw[arr] (task-b) -- (lm-b);
\draw[goldarr] (lm-b) -- (expert-b);
\node[note] at (5.4,-1.3) {Re-queries LM head every $N$ chunks};

\node[opbox] (op-c) at (10.8,0) {Operator subtask $\ell^{\text{sub}}$};
\node[expertbox] (expert-c) at (10.8,3.6) {Action expert\\+ flow head\\$\to A_t$};
\node[hdr] at (10.8,4.6) {(c) Interactive};
\draw[arr] (op-c) -- (expert-c);
\node[note] at (10.8,-1.3) {Holds operator's subtask across chunks until operator advances};

\end{tikzpicture}
\caption{Inference-time conditioning modes. The three panels show
where the action expert's low-level conditioning comes from. In \textbf{(a)
Flat} it is just the raw task string, re-fed every chunk with the LM head
never called. It is the only correct way to deploy the no-subtask arm (Run $1.4$
in \cref{tab:aloha-main}) since its LM head was never trained to produce
anything. \textbf{(b) Hierarchical} is how the trained FineART-VLA is deployed
(\cref{fig:architecture}): the LM head autoregressively generates a subtask
$\ell^{\text{sub}}$ every $N$ chunks and the action expert conditions on it
until the next regeneration, which is how the subtask-trained arm (Run
$1.3$) is deployed. \textbf{(c) Interactive} skips the LM head altogether and
lets an operator supply $\ell^{\text{sub}}$ directly. It is the
human-oracle protocol behind the long-horizon instruction-following result in
\cref{fig:subtask-heldout}, where the operator moves on to the next subtask
once the previous one is judged done rather than waiting on the LM head to
propose it.}
\label{fig:inference-modes}
\end{figure}
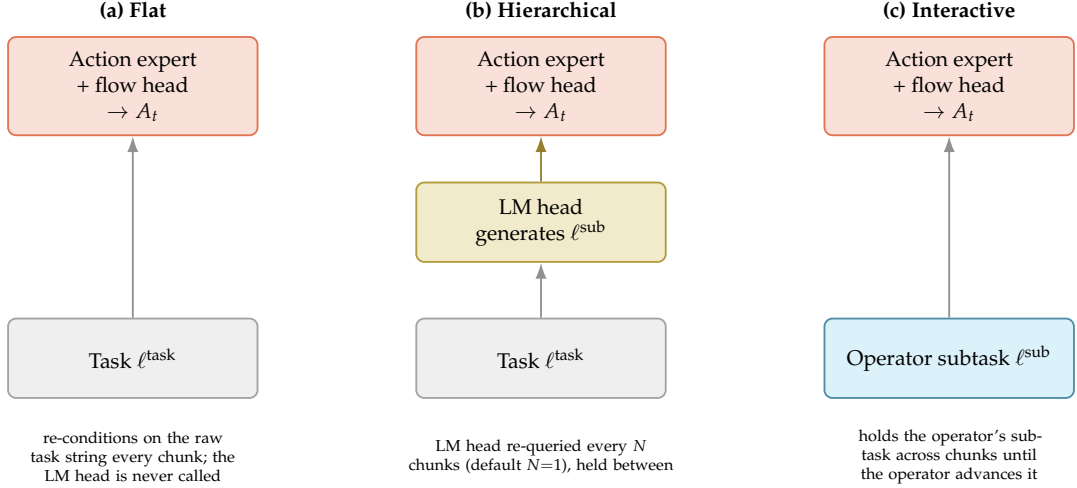

At deployment, the same checkpoint can be executed in three different ways,
independent of how it was mid-trained (\cref{fig:inference-modes}). All three
call the same action expert and flow-matching head from
\cref{fig:architecture}. They differ in where the low-level conditioning
comes from and how often it gets refreshed.

Chunk-level execution is the same in all three modes: a fresh action chunk is predicted from a new observation every \texttt{n\_action\_steps} control steps
while the policy runs open-loop in between. In \textbf{(b)} and \textbf{(c)}, the subtask
is held across several chunks so that it spans seconds rather than a single
$50$-step chunk and matches how subtasks are actually distributed in
training. We run inference eager rather than compiled because the action
expert's prompt embeds the discretized joint state as text. With token count shifting with almost every observation, a compiled graph re-specializes on nearly every call, so compilation never pays off.

\subsection{Language Runtime and Subtask Execution}
\label{sec:language_runtime}

FineART-VLA integrates with LeRobot's interactive rollout runtime,
which separates language-level instruction updates from action-chunk
execution. The runtime keeps the policy, observation processors, and
robot connection initialized throughout a session. A terminal interface
and a programmatic controller expose the same instruction-update and
execution operations.

\paragraph{Human-provided subtasks.}
An operator can replace the active instruction during execution without
reloading the model. Under synchronous inference, the runtime discards
queued policy actions computed under the previous instruction, and the
next prediction uses the updated instruction. When real-time chunking
is enabled, the new instruction conditions the next action chunk, which
is blended with the remaining action prefix.

\paragraph{Autonomous subtask inference.}
Given a high-level goal, the runtime periodically requests a subtask
from FineART-VLA using the current observation. The generation prompt
is constructed from the checkpoint's saved training recipe, preserving
the prompt structure used for subtask supervision. The generated text
becomes the active instruction for subsequent action predictions and
remains in effect until replaced. Each planning query receives the
original goal. The replanning interval is configurable and measured
from the application of the preceding subtask; the runtime does not
itself detect subtask completion.

\subsection{Efficient Training and Inference}
\label{sec:efficient_implementation}

\paragraph{Training.}
The implementation shares backbone computation across language,
FAST-token, and flow-matching objectives when their corresponding
supervision is present. To amortize visual-language prefix computation,
multiple independent noise and timestep draws reuse the same prefix,
with five draws by default. Action embeddings and output projections
are vectorized across these draws. Attention masks isolate the action
sequences and prevent access to FAST action targets, and the flow
losses are averaged across draws.

Additional optimizations include sparse and bucketed cross-entropy
computation, per-layer vision activation checkpointing, and fused
AdamW updates. Optional backends support compiled cross-entropy and
FlashRT~\cite{Su_FlashRT} adaptive RMSNorm kernels. For flow-only batches
with knowledge insulation and no language or FAST supervision, the
implementation can also omit the backbone's backward graph.

\paragraph{Inference.}
Autoregressive subtask generation uses incremental key--value caching.
The action-denoising implementation reuses the visual-language prefix
cache across steps and crops appended action entries instead of
copying the cache at each step. 

An optional FlashRT backend provides calibrated FP8 MLP kernels for
Gemma and SigLIP. These kernels are inference-only, change numerical
precision, and are disabled by default. 

Liang Su, author of FlashRT~\cite{Su_FlashRT}, contributed substantial implementation and optimization work underlying this section, including denoising-cache improvements, training-path optimizations, and FlashRT kernel integration.

\section{Evaluations}

Our evaluations total $3{,}400$ rollouts across $23$ checkpoints between experiments on ALOHA and YAM, summarized in \cref{tab:evals_overview}.

\begin{table}[H]
\centering
\caption{Summary of Evaluations}
\label{tab:evals_overview}
\small
\begin{tabularx}{\linewidth}{l c X X}
\toprule
\textbf{Task} & \textbf{Total Rollouts} & \textbf{ALOHA Experiment} & \textbf{YAM Experiment} \\
\midrule
Put cup on shelf & 575 & Benchmark & Benchmark \\
Fold towel & 575 & Benchmark & Benchmark \\
Insert flower in vase & 575 & Benchmark & Benchmark \\
Insert tool into pegboard & 575 & Benchmark & Benchmark \\
Put cup on saucer & 575 & Benchmark & Benchmark \\
Sort tools & 150 & Benchmark & Cross-Embodiment Transfer \\
Put donut into bin & 50 & Spatial Grounding & -- \\
Prepare breakfast & 125 & Long-Horizon & Cross-Embodiment Transfer \\
Put cable into bin & 200 & -- & Unseen Target and Distractors \\
\bottomrule
\end{tabularx}
\end{table}

\subsection{Procedure}

Evaluations of our scale demand procedural rigor. Across benchmark suites, we predetermine $25$ initial conditions, namely position and orientation, for each task (\cref{tab:benchmark_suites}).

\begin{table}[H]
  \centering
  \small
  \caption{Benchmark suites by initial conditions.}
  \label{tab:benchmark_suites}
  \begin{tabular}{lcc}
    \toprule
    \textbf{Benchmark Suite} & \textbf{Benchmark Tasks} & \textbf{Initial Conditions} \\
    \midrule
    In-Embodiment        & $6$ & $150$ \\
    Cross-Embodiment     & $5$ & $125$ \\
    \midrule
\textbf{Total}       & $\mathbf{11}$ & $\mathbf{275}$ \\
    \bottomrule
  \end{tabular}
\end{table}

\begin{figure}[b]
  \centering
  \includegraphics[width=\textwidth]{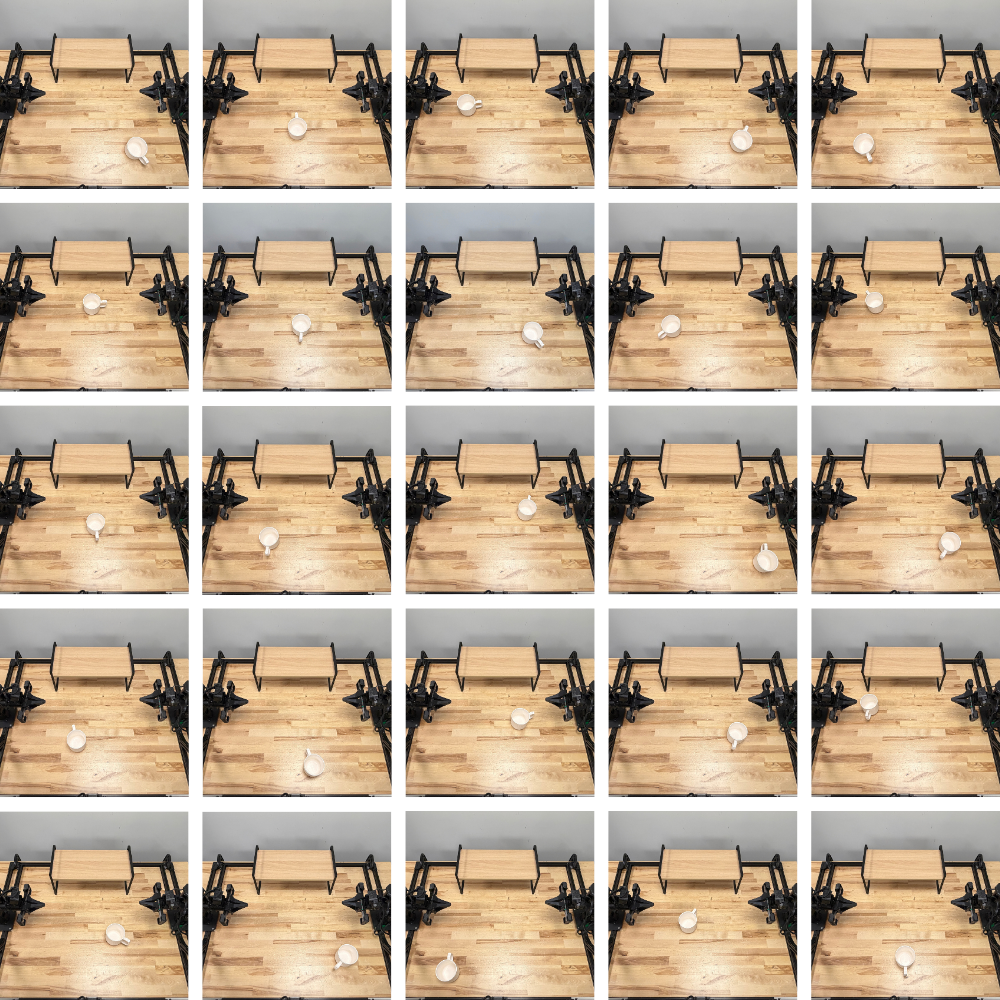}
  \caption{Initial conditions for put cup on shelf.}
  \label{fig:init_conditions}
\end{figure}

\cref{fig:init_conditions} illustrates the $25$ initial conditions for an in-distribution benchmark task: put cup on shelf.

\subsection{Rubric}

We evaluate each checkpoint by progress rate based on subtask completion percentage. Subtasks are defined such that one subtask must be completed before the next can begin (\cref{tab:tasks_by_subtask}). Prepare breakfast is an exception, with no prescribed order after its first subtask.

\begin{table}[htbp]
\centering
\small
\caption{Tasks by Subtask}
\label{tab:tasks_by_subtask}
\begin{tabularx}{\textwidth}{@{} l c X r @{}}
\toprule
\textbf{Task} & \textbf{Subtask \#} & \textbf{Subtask Description} & \textbf{Subtask Completion (\%)} \\
\midrule
\multirow{4}{*}{\textbf{Put cup on shelf}} 
 & 1 & Approach cup & 25.0 \\
 & 2 & Grasp cup & 50.0 \\
 & 3 & Move cup toward shelf & 75.0 \\
 & 4 & Place cup on shelf & 100.0 \\
\midrule
\multirow{8}{*}{\textbf{Fold towel}} 
 & 1 & Approach first boundary of towel & 12.5 \\
 & 2 & Grasp first boundary of towel & 25.0 \\
 & 3 & Move toward opposite boundary & 37.5 \\
 & 4 & Place at opposite boundary & 50.0 \\
 & 5 & Approach second boundary of towel & 62.5 \\
 & 6 & Grasp second boundary of towel & 75.0 \\
 & 7 & Move toward opposite boundary & 87.5 \\
 & 8 & Place at opposite boundary & 100.0 \\
\midrule
\multirow{4}{*}{\textbf{Insert flower in vase}} 
 & 1 & Approach flower & 25.0 \\
 & 2 & Grasp flower & 50.0 \\
 & 3 & Move flower toward vase & 75.0 \\
 & 4 & Insert flower in vase & 100.0 \\
\midrule
\multirow{4}{*}{\textbf{Insert tool into pegboard holder}} 
 & 1 & Approach tool & 25.0 \\
 & 2 & Grasp tool & 50.0 \\
 & 3 & Move tool toward pegboard & 75.0 \\
 & 4 & Insert tool into pegboard holder & 100.0 \\
\midrule
\multirow{4}{*}{\textbf{Put cup on saucer}} 
 & 1 & Approach cup & 25.0 \\
 & 2 & Grasp cup & 50.0 \\
 & 3 & Move cup toward saucer & 75.0 \\
 & 4 & Place cup on saucer & 100.0 \\
\midrule
\multirow{8}{*}{\textbf{Sort tools}} 
 & 1 & Approach first tool & 12.5 \\
 & 2 & Grasp first tool & 25.0 \\
 & 3 & Move toward bin & 37.5 \\
 & 4 & Drop object in corresponding bin & 50.0 \\
 & 5 & Approach second tool & 62.5 \\
 & 6 & Grasp second tool & 75.0 \\
 & 7 & Move toward bin & 87.5 \\
 & 8 & Drop object in corresponding bin & 100.0 \\
\midrule
\multirow{4}{*}{\textbf{Put donut into bin}} 
 & 1 & Approach donut & 25.0 \\
 & 2 & Grasp donut & 50.0 \\
 & 3 & Move donut toward correct bin & 75.0 \\
 & 4 & Place donut into correct bin & 100.0 \\
\midrule
\multirow{7}{*}{\textbf{Prepare breakfast}} 
 & 1 & Pick and place plate on tray & 14.3 \\
 & 2 & Pick up bread & 28.6 \\
 & 3 & Place bread on plate & 42.9 \\
 & 4 & Pick up pink donut & 57.1 \\
 & 5 & Place pink donut on plate & 71.4 \\
 & 6 & Pick up blue donut & 85.7 \\
 & 7 & Place blue donut on plate & 100.0 \\
\midrule
\multirow{2}{*}{\textbf{Put cable into bin}} 
 & 1 & Pick up cable & 50.0 \\
 & 2 & Put cable into bin & 100.0 \\
\bottomrule
\end{tabularx}
\end{table}

\clearpage
\section{Acknowledgments}
\label{app:ack}

Our work would not be possible without the support of colleagues and
friends at Scale AI, Hugging Face, and the community.

\begin{itemize}
\item We thank Alex Finch, Angel Uribe, Luke Pulaski, and Marci Ramos for their work in operations for data collection; Ana Paula Estevez, Juan Carlos Becerril, Jose Antonio Gonzalez, Paulina Vergara Montoya, and Astrid Hernandez Galvez for their work in operations for data annotation; Aidan Walker, Amanda Dee, Angel Uribe, Devender Bankoti, Luke Pulaski, and Marci Ramos for their work in operations for evaluation; and Aidan Walker, Aliyah Dela Cruz, Amanda Dee, Angel Uribe, Caroline Doubane, Devender Bankoti, Isaiah Smith, Joshua Diaz, Katerina Connearney, Marci Ramos, Reeya Shrestha, Rudy Garcia, Sherwood Yee, Tony Alfatooni, Travis Bringas, and Veronika Tsvelodub for their work as on-site contributors.

\item We thank Selam Gano, Sidney Nimako, Tyler Smithline, Martin Lombardo, Matias Del Carlo, Fran Espeche, Gabriel Blanchard, Shreyas Chakravarthula, Eric Taylor, and Tim Lu for their work in hardware engineering for data collection; Martin Lombardo, Matias Del Carlo, Fran Espeche, and Gianluca Bobbio for their work in software engineering for data collection; Gregorio Freidin, Matias Carou, Santiago Illi, and Gianluca Bobbio for their work in software engineering for data processing; and Garrett Matsuda, Juan Cabrera, Malena Goñi, Conrado Mader Blanco, Ezequiel Romio, Micaela Alvarez, Luis Lopez Castaneda, Ankit Vedak, Gianluca Bobbio, and Tim Lu for their work in software engineering for data annotation.

\item We thank Caroline Clark, Kendyl Burkitt, and Molly Taudvin for their go-to-market support; and Javier Gonzalez for their cross-functional support.

\item We thank Harsha Mohan, Michel Aractingi, Caroline Pascal, Carlos Jerez, Ke Wang, and Khalil Meftah for their early discussions on policy and dataset design.

\item We thank Liang Su for their work on training and inference speed optimizations for FineART-VLA; and Steven Palma and the LeRobot team for their contributions to the interactive rollout runtime.

\item We thank Ben Levin, Natasha Dadabhoy, and Joe Fox Jr. for their leadership and for approving the public release of the dataset.

\end{itemize}

\end{document}